\documentclass[sigconf, nonacm]{acmart}

\usepackage{amsmath,amsfonts}
\usepackage{graphicx}
\usepackage{textcomp}
\usepackage{xcolor}
\usepackage [english]{babel}
\usepackage [autostyle, english = american]{csquotes}
\MakeOuterQuote{"}
\usepackage{enumitem}
\usepackage{graphicx}
\usepackage{float}
\usepackage{dsfont}
\usepackage{graphics} 
\usepackage{epsfig} 
\usepackage{tabularx}
\usepackage{multicol}
\usepackage{multirow}
\usepackage{tabularx}
\usepackage{pdflscape}
\usepackage[noindentafter]{titlesec}
\usepackage{bbm}
\usepackage{bbding}
\usepackage{subfigure}
\usepackage{wasysym}
\usepackage{pifont}
\usepackage{afterpage}
\usepackage{capt-of}
\usepackage{bbm}
\usepackage{rotating}
\usepackage{lscape}
\usepackage{booktabs}
\usepackage{hyperref}
\hypersetup{bookmarks=false}
\usepackage{pgfplots}
\pgfplotsset{width=8cm,compat=1.9}
\graphicspath{ {images/} }
\usepackage{braket} 
\usepackage{algorithm}
\usepackage{xcolor}
\usepackage{mathtools}
\DeclarePairedDelimiter\norm{\lVert}{\rVert}
\usepackage[noend]{algpseudocode}
\renewcommand{\arraystretch}{1.1}
\newcommand{\blue}[1]{\textcolor{black}{#1}}
\newcommand{\red}[1]{\textcolor{black}{#1}}

\usepackage{mathtools,bm}

\newcommand\vldbdoi{XX.XX/XXX.XX}

\newcommand\vldbavailabilityurl{https://github.com/imperial-qore/TranAD}
 
\DeclareUnicodeCharacter{2212}{-}

\begin{document}
\urlstyle{tt}
\title{TranAD: Deep Transformer Networks for Anomaly Detection in Multivariate Time Series Data}

\author{Shreshth Tuli}
\affiliation{%
  \institution{Imperial College London}
  \city{London}
  \state{UK}
}
\email{s.tuli20@imperial.ac.uk}

\author{Giuliano Casale}
\affiliation{%
  \institution{Imperial College London}
  \city{London}
  \state{UK}
}
\email{g.casale@imperial.ac.uk}

\author{Nicholas R. Jennings}
\affiliation{%
  \institution{Loughborough University}
  \city{London}
  \state{UK}
}
\email{n.r.jennings@lboro.ac.uk}

\begin{abstract}
Efficient anomaly detection and diagnosis in multivariate time-series data is of great importance for modern industrial applications. However, building a system that is able to quickly and accurately pinpoint anomalous observations is a challenging problem. This is due to the lack of anomaly labels, high data volatility and the demands of ultra-low inference times in modern applications. Despite the recent developments of deep learning approaches for anomaly detection, only a few of them can address all of these challenges. In this paper, we propose TranAD, a deep transformer network based anomaly detection and diagnosis model which uses attention-based sequence encoders to swiftly perform inference with the knowledge of the broader temporal trends in the data. TranAD uses focus score-based self-conditioning to enable robust multi-modal feature extraction and adversarial training to gain stability. Additionally, model-agnostic meta learning (MAML) allows us to train the model using limited data. Extensive empirical studies on six publicly available datasets demonstrate that TranAD can outperform state-of-the-art baseline methods in detection and diagnosis performance with data and time-efficient training. Specifically, TranAD increases F1 scores by up to \blue{17\%}, reducing training times by up to 99\% compared to the baselines.
\end{abstract}

\maketitle


\vspace{-1pt}
\section{Introduction}
\noindent
Modern IT operations generate enormous amounts of \blue{high dimensional sensor data used for continuous monitoring and proper functioning of large-scale datasets. Traditionally, data mining experts have studied and highlighted data that do not follow usual trends to report faults. Such reports have been crucial for system management models for reactive fault tolerance and robust database design~\cite{tran2020real}.} However, with the advent of big-data analytics and deep learning, this problem has \blue{become of interest to data mining researchers and to aid experts} in handling increasing amounts of data. One particular use case is in artificial intelligence for Industry-4.0 databases, with a specific focus on service reliability~\cite{nedelkoski2020multi} that has automated fault detection, recovery and management of modern systems. Detecting \blue{data-faults, or any type of behavior not conforming to the expected trends,} is an active research discipline referred to as anomaly detection in multivariate time series~\cite{bulusu2020anomalous}. \blue{Many data-driven industries, including ones related to distributed computing, Internet of Things (IoT), robotics and urban resource management~\cite{audibert2020usad, thudumu2020comprehensive}} are now adopting machine learning based unsupervised methods for anomaly detection.

\textbf{Challenges.} \blue{The problem of anomaly detection is becoming increasingly challenging in large-scale databases due to the increasing data modality~\cite{he2021automl, witkowski2017internet, kingsbury2020elle}.} In particular, the increasing number of sensors and devices in contemporary IoT platforms with increasing data volatility creates the requirement for significant amounts of data for accurate inference. However, due to the rising federated learning paradigm with geographically distant clusters, \blue{synchronizing databases across devices is expensive, causing limited data availability for training~\cite{yang2019federated, tuli2021pregan}.} Further, next-generation applications \blue{need ultra-fast inference speeds for quick recovery and optimal Quality of Service (QoS)~\cite{bellendorf2020classification, tuli2021cosco, tuli2020modelling}. Time-series databases are generated using several engineering artifacts (servers, robots, \textit{etc.}) that interact with the environment, humans or other systems.} As a result, the data often displays both stochastic and temporal trends~\cite{omnianomaly}. It thus becomes crucial to distinguish outliers due to stochasticity and only pinpoint observations that do not adhere to the observed temporal trends. Moreover, the lack of labeled data and anomaly diversity makes the problem challenging as we \blue{cannot use supervised learning models, which have shown to be effective in other areas of data mining~\cite{chalapathy2019deep}.} Finally, it is not only important to detect anomalies but also the root causes\blue{, \textit{i.e.}, the specific data sources leading to abnormal behavior~\cite{jacob2020exathlon}.} This complicates the problem further as we need to perform multi-class prediction (whether there is an anomaly and from which source if so)~\cite{mscred}. 

\textbf{Existing solutions.} The above discussed challenges have led to the development of a myriad of unsupervised learning solutions for automated anomaly detection. Researchers have developed reconstruction-based methods that predominantly aim to encapsulate the temporal trends and predict the time-series data in an unsupervised fashion, then use the deviation of the prediction with the ground-truth data as anomaly scores. \blue{Based on various extreme value analysis methods, such approaches classify timestamps with high anomaly scores as abnormal~\cite{lstm_ndt, omnianomaly, audibert2020usad, mscred, gdn, mtad_gat, mad_gan, kingsbury2020elle, sand}.} The way prior works generate a predicted time-series from a given one varies from one work to another. \blue{Traditional approaches, like SAND~\cite{sand}, use clustering and statistical analysis to detect anomalies. Contemporary methods like openGauss~\cite{li2021opengauss} and LSTM-NDT~\cite{lstm_ndt} use a Long-Short-Term-Memory (LSTM) based neural networks to forecast the data with an input time-series and a non-parametric dynamic thresholding approach for detecting anomalies from prediction errors.} However, recurrent models like LSTMs are known to be slow and computationally expensive~\cite{audibert2020usad}. Recent state-of-the-art methods, like MTAD-GAT~\cite{mtad_gat} and \blue{GDN~\cite{gdn}, use deep neural networks with a time-series window as an input for more accurate predictions.} 
However, as the inputs become more data-intensive, small constant size window inputs limit the detection performance of such models due to the restricted local context information given to the model~\cite{audibert2020usad}. \blue{There is a need for a model that is fast and can capture high-level trends with minimal overheads.}

\textbf{New insights.} As noted above, recurrent models based on prior methods are not only slow and computationally expensive, but are also unable to model long-term trends effectively~\cite{gdn, mtad_gat, audibert2020usad}. This is because, at each timestamp, a recurrent model needs to first perform inference for all previous timestamps before proceeding further. Recent developments of the transformer models allow single-shot inference with the complete input series using position encoding~\cite{vaswani2017attention}. \blue{Using transformers allows much faster detection compared to recurrent methods by parallelizing inference on GPUs~\cite{huang2020hitanomaly}. However, transformers also provide the benefit of being able to encode large sequences with accuracy and training/inference times nearly agnostic to the sequence length~\cite{vaswani2017attention}.} Thus, we use transformers to grow the temporal context information sent to an anomaly detector without significantly increasing the computational overheads. 

\textbf{Our contributions.}  \blue{This work uses various tools, including Transformer neural networks and model-agnostic meta learning, as building blocks. However, each of these different technologies cannot be directly used and need necessary adaptations to create a generalizable model for anomaly detection. Specifically, we propose a transformer-based anomaly detection model (TranAD), that uses self-conditioning and an adversarial training process. Its architecture makes it fast for training and testing while maintaining stability with large input sequences. Simple transformer-based encoder-decoder networks tend to miss anomalies if the deviation is too small, \textit{i.e.}, it is relatively close to normal data. One of our contributions is to show that this can be alleviated by an adversarial training procedure that can amplify reconstruction errors. Further, using self-conditioning for robust multi-modal feature extraction can help gain training stability and allow generalization~\cite{liu2020diverse}. This, with model-agnostic meta learning (MAML) helps keep optimum detection performance even with limited data~\cite{finn2017model}, as we show later in the validation that methods with simple transformers underperform by over 11\% compared to TranAD.} We perform extensive empirical experiments on publicly available datasets to compare and analyze TranAD against the state-of-the-art methods. Our experiments show that TranAD is able to outperform baselines by increasing prediction scores by up to \blue{$17\%$} while reducing training time overheads by up to $99\%$.

The rest of the paper is organized as follows. Section~\ref{sec:related_work} overviews related work. Section~\ref{sec:method} outlines the working of the TranAD model for multivariate anomaly detection and diagnosis. A performance evaluation of the proposed method is shown in Section~\ref{sec:experiments}. Section~\ref{sec:analyses} presents additional analysis. Finally, Section~\ref{sec:conclusion} concludes.

\section{Related Work}
\label{sec:related_work}

Time series anomaly detection is a long-studied problem \blue{in the VLDB community}. The prior literature works on two types of time-series data: univariate and multivariate. For the former, various methods analyze and detect anomalies in time-series data with a single data source~\cite{mbiydzenyuy2020univariate}, while for the latter multiple time-series together~\cite{omnianomaly, gdn, mtad_gat}. 

\textbf{Classical methods.} \blue{Such methods for anomaly detection typically model the time-series distribution using various classical techniques like k-Mean clustering, Support Vector Machines (SVMs) or regression models~\cite{salem2014anomaly, ocsvm, sand, kingsbury2020elle}.} Other methods use wavelet theory or various signal transformation methods like Hilbert transform~\cite{kanarachos2015anomaly}. Other classes of methods use Principal Component Analysis (PCA), process regression or hidden Markov chains to model time-series data~\cite{patcha2007overview}. \blue{The GraphAn technique~\cite{boniol2020graphan} converts the time-series inputs to graphs and uses graph distance metrics to detect outliers.} Another technique, namely isolation forest, uses an ensemble of several isolation trees that recursively partition the feature space for outlier detection~\cite{liu2008isolation, if}. Finally, classical methods use variants of Auto-Regressive Integrated Moving Average (ARIMA) to model and detect anomalous behaviour~\cite{yaacob2010arima}. However, auto-regression based approaches are rarely used for anomaly detection in high-order multivariate time series due to their inability to efficiently capture volatile time-series~\cite{abbasimehr2020optimized}. \blue{Other methods like SAND~\cite{sand}, CPOD~\cite{tran2020real} and Elle~\cite{kingsbury2020elle} utilize clustering and database read-write history to detect outliers.}

Time-series discord discovery is another recently proposed method for fault prediction~\cite{yankov2008disk, yeh2016matrix, gharghabi2018matrix, merlin}. Time series discords refer to the most unusual time series subsequences, \textit{i.e.}, subsequences that are maximally different from all other subsequences in the same time series. A sub-class of methods uses matrix profiling or its variants for anomaly and motif discovery by detecting time series discords~\cite{monteiro2016stomp, gharghabi2018matrix, zhu2018matrix}. Many advances have been proposed to make matrix profiling techniques data and time efficient~\cite{imani2018matrix}. Other efforts aim to make matrix profiling applicable to diverse domains~\cite{zimmerman2019matrix}. However, matrix profiling has many more uses than just anomaly detection and is considered to be slower than pure discord discovery algorithms~\cite{merlin}. A recent approach, MERLIN~\cite{merlin}, uses a parameter-free version of time series discord discovery by iteratively comparing subsequences of varying length with their immediate neighbors. MERLIN is considered to be the state-of-the-art discord discovery approach with low overheads; hence, is regarded as one of the baselines in our experiments. 

\textbf{Deep Learning based methods.} Most contemporary state-of-the-art techniques employ some form of deep neural networks. The LSTM-NDT~\cite{lstm_ndt} method relies on an LSTM based deep neural network model that uses the input sequence as training data and, for each input timestamp, forecasts data for the next timestamp. \blue{LSTMs are auto-regressive neural networks that learn order dependence in sequential data, where the prediction at each timestamp uses feedback from the output of the previous timestamp.} This work also proposes a non-parametric dynamic error thresholding (NDT) strategy to set a threshold for anomaly labeling using moving averages of the error sequence. However, being a recurrent model, such models are slow to train in many cases with long input sequences. Further, LSTMs are often inefficient in modeling long temporal patterns, especially when the data is noisy~\cite{mtad_gat}. 

The DAGMM~\cite{dagmm} method uses a deep autoencoding Gaussian mixture model for dimension reduction in the feature space and recurrent networks for temporal modeling. This work predicts an output using a mixture of Gaussians, where the parameters of each Gaussian are given by a deep neural model. The autoencoder compresses an input datapoint into a latent space, that is then used by a recurrent estimation network to predict the next datapoint. The decoupled training of both networks allows the model to be more robust; however, it still is slow and unable to explicitly utilize inter-modal correlations~\cite{gdn}. The Omnianomaly~\cite{omnianomaly} uses a stochastic recurrent neural network (similar to an LSTM-Variational Autoencoder~\cite{lstm_vae}) and a planar normalizing flow to generate reconstruction probabilities. It also proposes an adjusted Peak Over Threshold (POT) method for automated anomaly threshold selection that outperforms the previously used NDT approach. This work led to a significant performance leap compared to the prior art, but at the expense of high training times. 

The Multi-Scale Convectional Recursive Encoder-Decoder (MSCRED)~\cite{mscred} converts an input sequence window into a normalized two-dimensional image and then passes it through a ConvLSTM layer. This method is able to capture more complex inter-modal correlations and temporal information, however is unable to generalize to settings with insufficient training data. The MAD-GAN~\cite{mad_gan} uses an LSTM based GAN model to model the time-series distribution using generators. This work uses not only the prediction error, but also the discriminator loss in the anomaly scores. MTAD-GAT~\cite{mtad_gat} uses a graph-attention network to model both feature and temporal correlations and pass it through a lightweight Gated-Recurrent-Unit (GRU) network that aids detection without severe overheads. \blue{Traditionally, attention operations perform input compression using convex combination where the weights are determined using neural networks. GRU is a simplified version of LSTM with a smaller parameter set and can be trained in limited data settings.} The CAE-M~\cite{cae_m} uses a convolutional autoencoding memory network, similar to MSCRED. It passes the time-series through a CNN with the output being processed by bidirectional LSTMs to capture long-term temporal trends. Such recurrent neural network-based models have been shown to have high computation costs and low scalability for high dimensional datasets~\cite{audibert2020usad}. 

More recent works such as USAD~\cite{audibert2020usad}, GDN~\cite{gdn} \blue{and openGauss~\cite{li2021opengauss} do not use resource-hungry recurrent models, but only attention-based network architectures to improve training speeds.} The USAD method uses an autoencoder with two decoders with an adversarial game-style training framework. This is one of the first works that focus on low overheads by using a simple autoencoder and can achieve a several-fold reduction in training times compared to the prior art. The Graph Deviation Network (GDN) approach learns a graph of relationships between data modes and uses attention-based forecasting and deviation scoring to output anomaly scores. \blue{The openGauss approach uses a tree-based LSTM that has lower memory and computational footprint and allows capturing temporal trends even with noisy data.} However, due to the small window as an input and the use of simple or no recurrent models, the latest models are unable to capture long-term dependencies effectively. 

The recently proposed HitAnomaly~\cite{huang2020hitanomaly} method uses vanilla transformers as encoder-decoder networks, but is only applicable to natural-language log data and not appropriate for generic continuous time-series data as inputs. In our experiments, we compare TranAD against the state-of-the-art methods MERLIN, LSTM-NDT, DAGMM, OmniAnomaly, MSCRED, MAD-GAN, USAD, MTAD-GAT, CAE-M and GDN. These methods have shown superiority in anomaly detection and diagnosis, but complement one another in terms of performance across different time-series datasets. Out of these, only USAD aims to reduce training times, but does this to a limited extent. Just like reconstruction based prior work~\cite{audibert2020usad, omnianomaly, mscred, mad_gan, cae_m}, we develop a TranAD model that learns broad level trends using training data to find anomalies in test data. We specifically improve anomaly detection and diagnosis performance with also reducing the training times in this work.

\section{Methodology}
\label{sec:method}

\begin{figure*}
    \centering 
    \includegraphics[width=\linewidth]{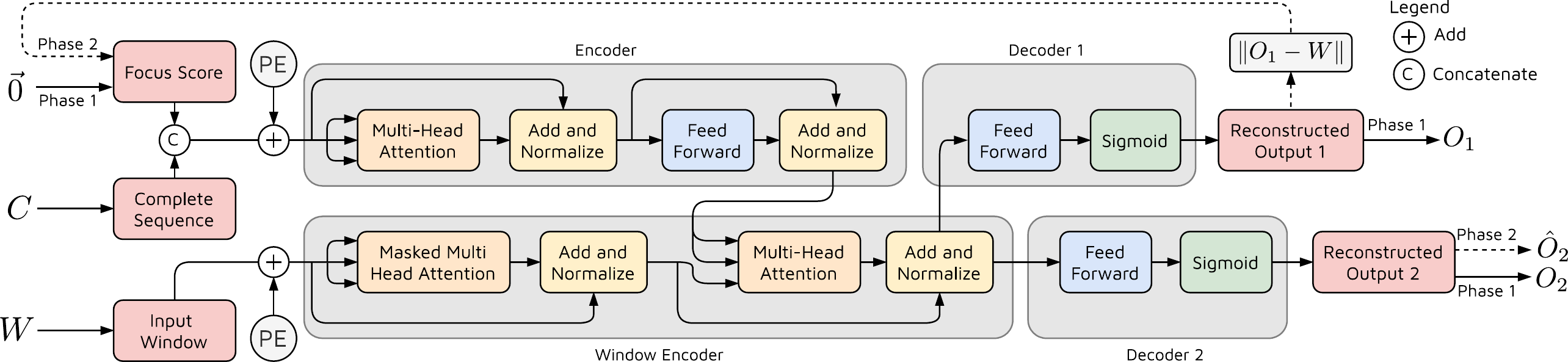}
    \caption{The TranAD Model.}
    \label{fig:model}
\end{figure*}

\subsection{Problem Formulation}
We consider a multivariate time-series, which is a timestamped sequence of observations/datapoints of size $T$
\[\mathcal{T} =\{x_1, \ldots, x_T\},\]
where each datapoint $x_t$ is collected at a specific timestamp $t$ and $x_t \in {\rm I\!R}^m,\ \forall t$. Here, the univariate setting is a particular case where $m = 1$. We now define the two problems of anomaly detection and diagnosis.

\textit{Anomaly Detection:} Given a training input time-series $\mathcal{T}$, for any unseen test time-series $\hat{\mathcal{T}}$ of length $\hat{T}$ and same modality as the training series, we need to predict $\mathcal{Y} = \{y_1, \ldots, y_{\hat{T}} \}$, where we use $y_t \in \{0, 1\}$ to denote whether the datapoint at the $t$-th timestamp of the test set is anomalous (1 denotes an anomalous datapoint). 

\textit{Anomaly Diagnosis:} Given the above training and test time-series, we need to predict $\mathcal{Y} = \{y_1, \ldots, y_{\hat{T}} \}$, where $y_t \in \{0, 1\}^m$ to denote which of the modes of the datapoint at the $t$-th timestamp are anomalous.

\subsection{Data Preprocessing}
To make our model more robust, we normalize the data and convert it to time-series windows, both for training and testing. We normalize the time-series as:
\begin{equation}
    x_t \gets \frac{x_t - \min(\mathcal{T})}{\max(\mathcal{T}) - \min(\mathcal{T}) + \epsilon'},
\end{equation}
where $\min(\mathcal{T})$ and $\max(\mathcal{T})$ are the mode wise minimum and maximum vectors in the training time-series. $\epsilon'$ is a small constant vector to prevent zero-division. \blue{Knowing the ranges a-priori, we normalize the data to get it in the range $[0, 1)$.}

To model the dependence of a data point $x_t$ at a timestamp $t$, we consider a local contextual window of length $K$ as
\[W_t = \{x_{t-K+1}, \ldots, x_t\}.\]
We use replication padding for $t < K$ and convert an input time series $\mathcal{T}$ to a sequence of sliding windows $\mathcal{W} = \{W_1, \ldots, W_T\}$. \blue{Replication padding, for each $t < K$, appends to the window $W_t$ a constant vector $\{x_t, \ldots, x_t\}$ of length $K-t$ to maintain the window length of $K$ for each $t$.} Instead of using $\mathcal{T}$ as training input, we use $\mathcal{W}$ for model training and $\hat{\mathcal{W}}$ (corresponding to $\hat{\mathcal{T}}$) as the test series. This is a common practice in prior work~\cite{audibert2020usad, omnianomaly} as it allows us to give a datapoint with its local context instead of a standalone vector, and hence is used in our model. We also consider the time slice until the current timestamp $t$ of a series $\mathcal{T}$ and denote it as $C_t$. 

Now, instead of directly predicting the anomaly label $y_t$ for each input window $W_t$, we shall first predict an anomaly score $s_t$ for this window. Using anomaly scores for the past input windows, we calculate a threshold value $D$, above which we label the input window as anomalous, thus $y_t = \mathds{1}(s_t \geq D)$. To calculate the anomaly score $s_t$, we reconstruct the input window as $O_t$ and use the deviation between $W_t$ and $O_t$. For the sake of simplicity and without loss of generality, we shall use $W$, $C$, $O$ and $s$ for the rest of the discussion.

\subsection{Transformer Model}
Transformers are popular deep learning models that have been used in various natural language and vision processing tasks~\cite{vaswani2017attention}. However, we use insightful refactoring of the transformer architecture for the task of anomaly detection in time-series data. Just like other encoder-decoder models, in a transformer, an input sequence undergoes several attention-based transformations. Figure~\ref{fig:model} shows the architecture of the neural network used in TranAD. The encoder encodes the complete sequence until the current timestamp $C$ with a focus score (more details later). The window encoder uses this to create an encoded representation of the input window $W$, which is then passed to two decoders to create its reconstruction. 

We now provide details on the working of TranAD. A multivariate sequence like $W$ or $C$ is transformed first into a matrix form with modality $m$. We define scaled-dot product attention~\cite{vaswani2017attention} of three matrices $Q$ (query), $K$ (key) and $V$ (value):
\begin{equation}
    \mathrm{Attention}(Q, K, V) = \mathrm{softmax}\left(\frac{QK^T}{\sqrt{m}}\right)V.
\end{equation}
\blue{Here, the $\mathrm{softmax}$ forms the convex combination weights for the values in $V$, allowing us to compress the matrix $V$ into a smaller representative embedding for simplified inference in the downstream neural network operations. Unlike traditional attention operation, the scaled-dot product attention scales the weights by a $\sqrt{m}$ term to reduce the variance of the weights, facilitating stable training~\cite{vaswani2017attention}.} For input matrices $Q$, $K$ and $V$, we apply Multi-Head Self Attention~\cite{vaswani2017attention} by first passing it through $h$ (number of heads) feed-forward layers to get $Q_i$, $K_i$ and $V_i$ for $i \in \{1, \ldots, h\}$, and then applying scaled-dot product attention as
\begin{align}
\begin{split}
    \mathrm{MultiHeadAtt}(Q, K, V) &= \mathrm{Concat}(H_1, \ldots, H_h),\\
    \text{where } H_i &= \mathrm{Attention}(Q_i, K_i, V_i).
\end{split}
\end{align}
Multi-Head Attention allows the model to jointly attend to information from different representation sub-spaces at different positions. In addition, we use position encoding of the input matrices as defined in~\cite{vaswani2017attention}.

As GAN models have been shown to perform well in characteristic tasks of whether an input is anomalous or not, we leverage a time-efficient GAN style adversarial training method. Our model consists of two transformer encoders and two decoders (Figure~\ref{fig:model}). We consider the model inference in two phases. We first take the $W$ and $C$ pair as an input and a focus score $F$ (initially a zero matrix of the dimension of $W$, more details in the next subsection). We broadcast $F$ to match the dimension of $W$, with appropriate zero-padding and concatenate the two. We then apply position encoding and obtain the input for the first encoder, say $I_1$. The first encoder performs the following operations 
\begin{align}
\begin{split}
\label{eq:encoder}
    I_1^1 &= \mathrm{LayerNorm}(I_1 + \mathrm{MultiHeadAtt}(I_1, I_1, I_1)),\\
    I_1^2 &= \mathrm{LayerNorm}(I_1^1 + \mathrm{FeedForward}(I_1^1)).
\end{split}
\end{align}
Here, $\mathrm{MultiHeadAtt}(I_1, I_1, I_1)$ denotes the multi-head self attention operation for the input matrix $I_1$ and $+$ denotes matrix addition. \blue{The above operations generate attention weights using the input time-series windows and the complete sequence to capture temporal trends within the input sequences. These operations enable the model to infer over multiple batches of the time-series windows in parallel as the neural network, at each timestamp, does not depend on the output of a previous timestamp, significantly improving the training time of the proposed method.} For the window encoder, we apply position encoding to the input window $W$ to get $I_2$. We modify the self-attention in the \blue{window encoder} to mask the data at subsequent positions. This is done to prevent the decoder from looking at the datapoints for future timestamp values at the time of training as all data $W$ and $C$ is given at once to allow parallel training. The window encoder performs the following operations
\begin{align}
\begin{split}
\label{eq:window_encoder}
    I_2^1 &= \mathrm{Mask}(\mathrm{MultiHeadAtt}(I_2, I_2, I_2)), \\
    I_2^2 &= \mathrm{LayerNorm}(I_2 + I_2^1),\\
    I_2^3 &= \mathrm{LayerNorm}(I_2^2 + \mathrm{MultiHeadAtt}(I_1^2, I_1^2, I_2^2)).
\end{split}
\end{align}
The encoding of the complete sequence $I_1^2$ is used as value and keys by the window encoder for the attention operation using the encoded input window as the query matrix. \blue{The motivation behind the operations in~\eqref{eq:window_encoder} is similar to the one for~\eqref{eq:encoder}; however, here we apply masking of the window input to hide the window sequences for future timestamps in the same input batch.} As the complete input sequence up to the $t$-th timestamp is given to the model as an input; it allows the model to encapsulate and leverage a larger context compared to a bounded, limited one as in prior art~\cite{audibert2020usad, omnianomaly, mtad_gat}.    
Finally, we use two identical decoders which perform the operation
\begin{equation}
    \label{eq:decoder}
    O_i = \mathrm{Sigmoid}(\mathrm{FeedForward}(I_2^3)),
\end{equation}
where $i \in \{1, 2\}$ for the first and second decoder respectively. The $\mathrm{Sigmoid}$ activation is used to generate an output in the range $[0,1]$, to match the normalized input window. Thus, the TranAD model takes the inputs $C$ and $W$ to generate two outputs $O_1$ and $O_2$. 

\subsection{Offline Two-Phase Adversarial Training}
We now describe the adversarial training process and the two-phase inference approach in the TranAD model, summarized in Algorithm~\ref{alg:training}. 

\begin{algorithm}[t]
    \begin{algorithmic}[1]
    \Require
    \Statex Encoder $E$, Decoders $D_1$ and $D_2$
    \Statex Dataset used for training $\mathcal{W}$
    \Statex Evolutionary hyperparameter $\epsilon$
    \Statex Iteration limit $N$
    \State Initialize weights $E, D_1, D_2$
    \State $n \gets 0$
    \State \textbf{do}
    \State \hspace{8pt} \textbf{for}($t = 1 \text{ to } T$)
    \State \hspace{15pt} $O_1, O_2 \gets D_1(E(W_t, \vec{0})), D_2(E(W_t, \vec{0}))$
    \State \hspace{15pt} $\hat{O}_2 \gets D_2(E(W_t, \norm{O_1 - W_t}_2))$
    \State \hspace{15pt} $L_1 = \epsilon^{-n} \norm{O_1 - W_t}_2 + (1 - \epsilon^{-n}) \norm{\hat{O}_2 - W_t}_2$ \label{line:l1}
    \State \hspace{15pt} $L_2 = \epsilon^{-n} \norm{O_2 - W_t}_2 - (1 - \epsilon^{-n}) \norm{\hat{O}_2 - W_t}_2$ \label{line:l2}
    \State \hspace{15pt} Update weights of $E, D_1, D_2$ using $L_1, L_2$ 
    \State \hspace{15pt} $n \gets n + 1$
    \State \hspace{8pt} Meta-Learn weights $E, D_1, D_2$ using a random batch \label{line:meta}
    \State \textbf{while} $n < N$
    \end{algorithmic}
\caption{The TranAD training algorithm}
\label{alg:training}
\end{algorithm}

\textbf{Phase 1 - Input Reconstruction.} \blue{The Transformer model enables us to predict the reconstruction of each input time-series window. It does this by acting as an encoder-decoder network at each timestamp. However, traditional encoder-decoder models often are unable to capture short-term trends and tend to miss anomalies if the deviations are too small~\cite{mad_gan}. To tackle this challenge, we develop an auto-regressive inference style that predicts the reconstructed window in two-phases. In the first phase, the model aims to generate an approximate reconstruction of the input window. The deviation from this inference, referred to as the \textit{focus score} mentioned previously, facilitates the attention network inside the Transformer Encoder to extract temporal trends, focusing on the sub-sequences where the deviations are high. Thus, the output of the second phase is conditioned on the deviations generated from the first phase.} Thus, in the first stage, the encoders convert the input window $W \in {\rm I\!R}^{K\times m}$ (with focus score $F=[0]_{K\times m}$) to a compressed latent representation $I_2^3$ using context-based attention as in a common transformer model. This compressed representation is then converted to generate outputs $O_1$ and $O_2$ via Eq.~\eqref{eq:decoder}. 

\textbf{Phase 2 - Focused Input Reconstruction.} \blue{In the second phase, we use the reconstruction loss for the first decoder as a focus score.} Having the focus matrix for the second phase $F = L_1$, we rerun model inference to obtain the output of the second decoder as $\hat{O}_2$.  

\blue{The focus score generated in the first phase indicates the deviations of the reconstructed output from the given input. This acts as a prior to modify the attention weights in the second phase and gives higher neural network activation to specific input sub-sequences to extract short-term temporal trends. We refer to this approach as "self-conditioning" in the rest of the paper. This two-phase auto-regressive inference style has a three-fold benefit. First, it amplifies the deviations, as the reconstruction error acts as an activation in the attention part of the Encoder in Figure~\ref{fig:model}, to generate an anomaly score, simplifying the fault-labeling task (discussed in Section~\ref{sec:online_inference}). Second, it prevents false positives by capturing short-term temporal trends in the Window Encoder in Figure~\ref{fig:model}. Third, the adversarial style training is known to improve generalizability and make the model robust to diverse input sequences~\cite{audibert2020usad}.}

\blue{\textbf{Evolving Training Objective.} The above-described model is bound to suffer from similar challenges as in other adversarial training frameworks. One of the critical challenges is maintaining training stability. To tackle this, we design an adversarial training procedure that uses outputs from two separate decoders (Decoders 1 and 2 in Figure~\ref{fig:model}). Initially, both decoders aim to independently reconstruct the input time-series window. } As in~\cite{omnianomaly} and~\cite{lstm_vae}, we define the reconstruction loss for each decoder using the L2-norm \blue{using the outputs of the first phase}:
\begin{align}
\begin{split}
    L_1 &= \norm{O_1 - W}_2,\\
    L_2 &= \norm{O_2 - W}_2.
\end{split}
\end{align} 
\blue{We now introduce the adversarial loss that uses outputs of the second phase.} Here, the second decoder aims to distinguish between the input window and the candidate reconstruction generated by the first decoder in phase 1 (using the focus scores) by maximizing the difference $||\hat{O}_2 - W||_2$. On the other hand, the first decoder aims to fool the second decoder by aiming to create a degenerate focus score (a zero vector) by perfectly reconstructing the input (\textit{i.e.}, $O_1 = W$). This pushes the decoder 2, in this phase, to generate the same output as $O_2$ which it aims to match the input in phase 1. This means the training objective is
\begin{equation}
    \min_{\mathrm{Decoder 1}} \max_{\mathrm{Decoder 2}} \norm{\hat{O}_2 - W}_2.
\end{equation}
Thus, the objective of the first decoder is to minimize the reconstruction error of this self-conditioned output, whereas the objective of the second one is to maximize the same. We realize this by using the loss as:
\begin{align}
\begin{split}
    L_1 &= + \norm{\hat{O}_2 - W}_2,\\
    L_2 &= - \norm{\hat{O}_2 - W}_2.
\end{split}
\end{align}

Now that we have loss functions for both phases, we need to determine the cumulative loss for each decoder. We thus use an evolutionary loss function that combines the \blue{reconstruction and adversarial} loss functions from the two phases as
\begin{align}
\begin{split}
    L_1 &= \epsilon^{-n} \norm{O_1 - W}_2 + (1 - \epsilon^{-n}) \norm{\hat{O}_2 - W}_2,\\
    L_2 &= \epsilon^{-n} \norm{O_2 - W}_2 - (1 - \epsilon^{-n}) \norm{\hat{O}_2 - W}_2,
\end{split}
\end{align}
where $n$ is the training epoch and $\epsilon$ is a training parameter close to one (lines \ref{line:l1}-\ref{line:l2} in Alg.~\ref{alg:training}). \blue{Initially, the weight given to the reconstruction loss is high. This is to ensure stable training when the outputs of the decoders are poor reconstructions of the input window. With poor reconstructions, the focus scores used in the second phase would be unreliable; and hence, cannot be utilized as a prior to indicating reconstructions that are far from the input sequence. Thus, the adversarial loss is given a low weight in the initial part of the process to avoid destabilizing model training. As reconstructions become closer to the input windows, and focus scores become more precise, the weight to the adversarial loss is increased. As loss curves in the neural network training process typically follow exponential function, we use weights of the form $\epsilon^{-n}$ in the training process with a small positive constant parameter $\epsilon$.}

As the training process does not assume that the data is available sequentially (as in an online process), the complete time-series can be split into ($W$, $C$) pairs and the model can be trained using input batches. \blue{Masked multi-head attention allows us to run this in parallel across several batches and speed up the training process.}

\begin{algorithm}[t]
    \begin{algorithmic}[1]
    \Require
    \Statex Trained Encoder $E$, Decoders $D_1$ and $D_2$
    \Statex Test Dataset $\hat{\mathcal{W}}$
    \State \textbf{for}($t = 1 \text{ to } \hat{T}$)
    \State \hspace{8pt} $O_1, O_2 \gets D_1(E(\hat{W}_t, \vec{0})), D_2(E(\hat{W}_t, \vec{0}))$ \label{line:inference}
    \State \hspace{8pt} $\hat{O}_2 \gets D_1(E(\hat{W}_t, \norm{O_1 - W}_2)), D_2(E(\hat{W}_t, \norm{O_1 - W}_2))$ \label{line:inference2}
    \State \hspace{8pt} $s = \tfrac{1}{2} \norm{O_1 - \hat{W}}_2 + \tfrac{1}{2} \norm{\hat{O}_2 - \hat{W}}_2$ \label{line:ascore}
    \State \hspace{8pt} $y_i = \mathds{1}(s_i \geq \mathrm{POT}(s_i))$ \label{line:diagnosis}
    \State \hspace{8pt} $y = \underset{i}{\lor} y_i$ \label{line:detection}
    \end{algorithmic}
\caption{The TranAD testing algorithm}
\label{alg:testing}
\end{algorithm}

\textbf{Meta Learning.} Finally, our training loop uses model-agnostic meta learning (MAML), a few-shot learning model for fast adaptation of neural networks~\cite{finn2017model}. This helps our TranAD model learn temporal trends in the input training time-series with limited data. In each training epoch, a gradient update for neural network weights (without loss in generality assume $\theta$) can be simply written as 
\begin{equation}
    \theta' \gets \theta - \alpha \nabla_\theta L(f(\theta)),
\end{equation}
where $\alpha$, $f(\cdot)$ and $L(\cdot)$ are learning rate, abstract representation of the neural network and loss function respectively. Now, at the end of each epoch we perform meta-learning step as
\begin{equation}
    \theta \gets \theta - \beta \nabla_\theta L(f(\theta')).
\end{equation}
The meta-optimization is performed with a meta step-size $\beta$, over the model weights $\theta$ where the objective is evaluated using the updated weights $\theta'$. Prior work has shown that this allows models to be trained quickly with limited data~\cite{finn2017model}. We encapsulate this in a single line in Algorithm~\ref{alg:training} (line~\ref{line:meta}).

\subsection{Online Inference, Anomaly Detection and Diagnosis}
\label{sec:online_inference}
\begin{figure}[]
    \centering \setlength{\belowcaptionskip}{-10pt}
    \includegraphics[width=\linewidth]{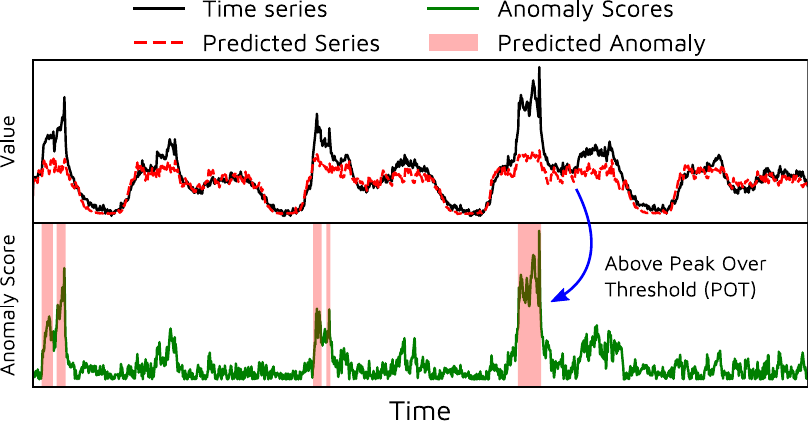}
    \caption{Visualization of anomaly prediction.}
    \label{fig:test}
\end{figure}

We now describe the inference procedure using the trained transformer model (summarized in Algorithm~\ref{alg:testing}). For an unseen data ($\hat{W}$, $\hat{C}$), the anomaly score is defined as
\begin{equation}
    s = \tfrac{1}{2} \norm{O_1 - \hat{W}}_2 + \tfrac{1}{2} \norm{\hat{O}_2 - \hat{W}}_2.
\end{equation}
The inference at test time runs again in two phases and hence we get a single pair of reconstruction ($O_1$, $\hat{O}_2$) (\blue{lines \ref{line:inference} and~\ref{line:inference2}} in Alg.~\ref{alg:testing}). At test time, we only consider the data until the current timestamp and hence this operation runs sequentially in an online fashion. Once we have the anomaly scores for a timestamp for each dimension $s_i$, we label the timestamp anomalous if this score is greater than a threshold. \blue{As is common in prior work~\cite{omnianomaly, lstm_ndt, boniol2020graphan}, for fair comparison,} we use the Peak Over Threshold (POT)~\cite{siffer2017anomaly} method to choose the threshold automatically and dynamically. \blue{In essence, this is a statistical method that uses ``extreme value theory'' to fit the data distribution with a Generalized Pareto Distribution and identify appropriate value at risk to dynamically determine threshold values. We also tested with another popular EVT method, namely annual maximum (AM)~\cite{bezak2014comparison}; however, we have observed 7.2\% higher F1 scores on an average for TranAD with POT than AM.} Anomaly diagnosis label for each dimension $i$ ($y_i$) and detection ($y$) results is defined as
\begin{align}
\begin{split}
    y_i &= \mathds{1}(s_i \geq \mathrm{POT}(s_i)),\\
    y &= \underset{i}{\lor} y_i.
\end{split}
\end{align}
Thus, we label the current timestamp anomalous if any of the $m$ dimensions is anomalous (lines \ref{line:diagnosis}-\ref{line:detection} in Alg.~\ref{alg:testing}). Figure~\ref{fig:test} illustrates this process for a sample time-series.

\begin{figure}[]
    \centering 
    \includegraphics[width=\linewidth]{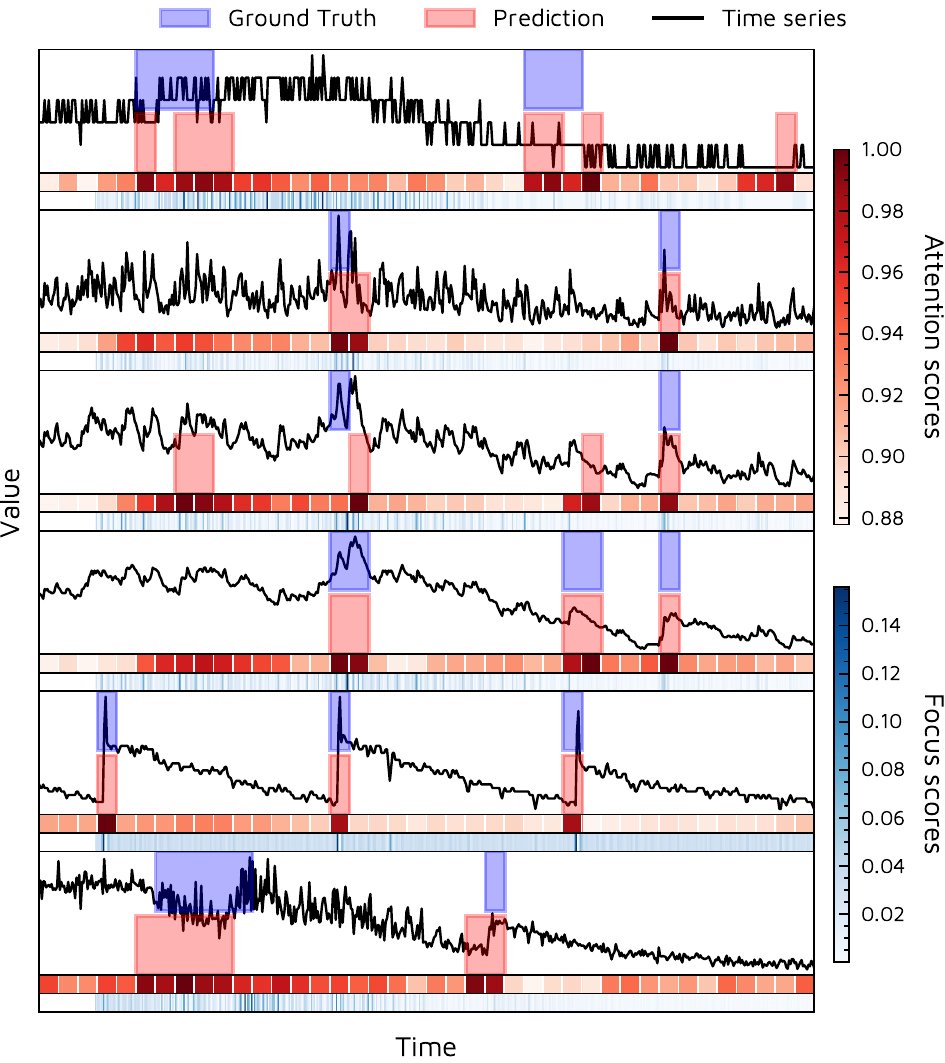}
    \caption{Visualization of focus and attention scores.}
    \label{fig:visualization}
\end{figure}

\textbf{Impact of Attention and Focus Scores.} Figure~\ref{fig:visualization} visualizes the attention and focus scores for the TranAD model trained on the SMD dataset (details in Section~\ref{sec:datasets}). We show the time-series, the average attention weights for each window (averaged over multiple heads) and focus scores for the first six dimensions of the dataset. It is apparent that the focus scores are highly correlated with the peaks and noise in the data. There is also a high correlation of focus scores across dimensions. For timestamps with sudden changes in the time-series, focus scores are higher. Further, the model gives higher attention weights to the specific dimensions of the time-series where the deviations are higher. This allows the model to specifically detect anomalies in each dimension individually, with the contextual trend of the complete sequence as a prior.

\section{Experiments}
\label{sec:experiments}
\noindent
We compare TranAD with state-of-the-art models for mutlivariate time-series anomaly detection, including MERLIN~\cite{merlin}, LSTM-NDT~\cite{lstm_ndt} \blue{(with autoencoder implementation from openGauss~\cite{li2021opengauss})}, DAGMM~\cite{dagmm}, OmniAnomaly~\cite{omnianomaly}, MSCRED~\cite{mscred}, MAD-GAN~\cite{mad_gan}, USAD~\cite{audibert2020usad}, MTAD-GAT~\cite{mtad_gat}, CAE-M~\cite{cae_m} and GDN~\cite{gdn} \blue{(with graph embedding implementation from GraphAn~\cite{boniol2020graphan})} . For more details refer Section~\ref{sec:related_work}.\footnote{We use publicly available code sources for most of the baselines. LSTM-NDT~\url{https://github.com/khundman/telemanom}, \blue{openGauss~\url{https://gitee.com/opengauss/openGauss-AI},} DAGMM \url{https://github.com/tnakae/DAGMM}, OmniAnomaly \url{https://github.com/NetManAIOps/OmniAnomaly}, MSCRED \url{https://github.com/7fantasysz/MSCRED}, MAD-GAN \url{https://github.com/LiDan456/MAD-GANs}. Other models were re-implemented by us. \red{See Appendix~\ref{app:merlin} for implementation of MERLIN baseline.}} We also tested the Isolation Forest method, but due to its low F1 scores, do not include the corresponding results in our discussion. Other classical methods have been omitted as deep-learning based approaches have already been shown to outperform them in prior work~\cite{mtad_gat, gdn, audibert2020usad}. We use hyperparameters of the baseline models as presented in their respective papers. We train all models using PyTorch-1.7.1~\cite{paszke2019pytorch} library\footnote{\blue{Parallel Transformer training was implemented as per~\cite{vaswani2017attention}.} All model training and experiments were performed on a system with configuration: Intel i7-10700K CPU, 64GB RAM, Nvidia RTX 3080 and Windows 11 OS.}. 

We use the AdamW~\cite{kingma2014adam} optimizer to train our model with an initial learning rate of $0.01$ (meta learning rate $0.02$) and step-scheduler with step size of $0.5$~\cite{saleh2019dynamic}. We use the following hyperparameter values.
\begin{itemize}[leftmargin=*]
    \item Window size = 10. 
    \item Number of layers in transformer encoders = 1
    \item Number of layers in feed-forward unit of encoders = 2
    \item Hidden units in encoder layers = 64
    \item Dropout in encoders = 0.1
\end{itemize}
The effect of window size on anomaly detection performance is analyzed in Section~\ref{sec:analyses}. \blue{We choose hyperparameters other than the window size using grid search.} For POT parameters, $\mathrm{coefficient} = 10^{−4}$ for all data sets, low quantile is 0.07 for SMAP, 0.01 for MSL, and 0.001 for others. These were selected as per the implementation of the OmniAnomaly baseline~\cite{omnianomaly}.
The only dataset-specific hyperparameter is the number of heads in multi-head attention, which was kept to be the same as the dimension size of the dataset. Other assignments for this hyperparameter give similar broad-level trends.

\blue{To train TranAD, we divide the training time-series into $80\%$ training data and $20\%$ validation data. To avoid model over-fitting, we use early-stopping criteria to train TranAD, \textit{i.e.}, we stop the training process once the validation accuracy starts to decrease.}
 
\begin{table}[t]
    \centering  
    \caption{Dataset Statistics}
    \resizebox{\linewidth}{!}{
    \begin{tabular}{@{}llrrrr@{}}
    \toprule
    Dataset &  & Train & Test & Dimensions & Anomalies (\%)\tabularnewline
    \midrule
    \blue{NAB} &  & \blue{4033} & \blue{4033} & \blue{1 (6)} & \blue{0.92}\tabularnewline
    UCR &  & 1600 & 5900 & 1 (4) & 1.88\tabularnewline
    MBA &  & 100000 & 100000 & 2 (8) & 0.14\tabularnewline
    SMAP &  & 135183 & 427617 & 25 (55) & 13.13\tabularnewline
    MSL &  & 58317 & 73729 & 55 (3) & 10.72\tabularnewline
    SWaT &  & 496800 & 449919 & 51 (1) & 11.98\tabularnewline
    WADI &  & 1048571 & 172801 & 123 (1) & 5.99\tabularnewline
    SMD &  & 708405 & 708420 & 38 (4) & 4.16\tabularnewline
    MSDS &  & 146430 & 146430 & 10 (1) & 5.37\tabularnewline
    \bottomrule
    \end{tabular}}
    \label{tab:datasets}
\end{table} 
\raggedbottom
 
\subsection{Datasets}
\label{sec:datasets}
We use seven publicly available datasets in our experiments. We summarize their characteristics in Table~\ref{tab:datasets}. The values in parenthesis are the number of sequences in the dataset repository and we report average scores across all sequences in a dataset \red{(NAB and UCR are univariate, the remaining ones are multivariate datasets)}. For instance, the SMAP dataset has 55 traces with 25 dimensions each. \red{While we share some of the concerns expressed in~\cite{wu2020current} about the lack of quality benchmark datasets for time series anomaly detection, we use these commonly-used benchmark datasets here to enable direct comparison of our approach to competing methods.}
\begin{enumerate}[leftmargin=*]
    \item \blue{\textit{Numenta Anomaly Benchmark (NAB)}: is a dataset of multiple real-world data traces, including readings from temperature sensors, CPU utilization of cloud machines, service request latencies and taxi demands in New York city~\cite{ahmad2017unsupervised}. However, this dataset is known to have sequences with incorrect anomaly labels such as the \texttt{nyc-taxi} trace\red{~\cite{merlin}}, which we exclude in our experiments.}
    \item \textit{HexagonML (UCR) dataset}: is a dataset of multiple univariate time series (included just for completeness) that was used in KDD 2021 cup~\cite{ucr, UCRArchive2018}. We include only the datasets obtained from natural sources (the \texttt{InternalBleeding} and \texttt{ECG} datasets) and ignore the synthetic sequences.
    \item \textit{MIT-BIH Supraventricular Arrhythmia Database (MBA)}: is a collection of electrocardiogram recordings from four patients, containing multiple instances of two different kinds of anomalies (either supraventricular contractions or premature heartbeats)~\cite{goldberger2000physiobank, moody2001impact}. This is a popular large-scale dataset in the data management community~\cite{boniol2020automated, sand}.
    \item \textit{Soil Moisture Active Passive (SMAP) dataset}: is a dataset of soil samples and telemetry information using the Mars rover by NASA~\cite{lstm_ndt}.
    \item \textit{Mars Science Laboratory (MSL) dataset}: is a dataset similar to SMAP but corresponds to the sensor and actuator data for the Mars rover itself~\cite{lstm_ndt}. However, this dataset is known to have many trivial sequences; hence, we consider only the three non-trivial ones (\texttt{A4}, \texttt{C2} and \texttt{T1}) pointed out by~\cite{merlin}.
    \item \textit{Secure Water Treatment (SWaT) dataset}: This dataset is collected from a real-world water treatment plant with 7 days of normal and 4 days of abnormal operation~\cite{mathur2016swat}. This dataset consists of sensor values (water level, flow rate, etc.) and actuator operations (valves and pumps).
    \item \textit{Water Distribution (WADI) dataset}: This is an extension of the SWaT system, but had more than twice the number of sensors and actuators than the SWaT model~\cite{ahmed2017wadi}. The dataset is also collected for a longer duration of 14 and 2 days of normal and attack scenarios. 
    \item \textit{Server Machine Dataset (SMD)}: This is a five-week long dataset of \red{stacked traces of the} resource utilizations of 28 machines from a compute cluster~\cite{omnianomaly}. Similar to MSL, we use the non-trivial sequences in this dataset, specifically the traces named \texttt{machine-1-1}, \texttt{2-1}, \texttt{3-2} and \texttt{3-7}.
    \item \textit{Multi-Source Distributed System (MSDS) Dataset}: This is a recent high-quality multi-source data composed of distributed traces, application logs, and metrics from a complex distributed system~\cite{nedelkoski2020multi}. This dataset is specifically built for AI operations, including automated anomaly detection, root cause analysis, and remediation.
\end{enumerate}
\blue{We eschew comparisons on the Yahoo~\cite{yahoo} dataset that has been claimed to suffer from mislabeling and run-to-failure bias\red{~\cite{wu2020current}}.}

\begin{table*}[!t]
    \centering 
    \caption{Performance comparison of TranAD with baseline methods on the complete dataset.  P: Precision, R: Recall, AUC: Area under the ROC curve, F1: F1 score with complete training data. The best F1 and AUC scores are highlighted in bold.} 
    \begin{tabular}{llcccccccccccccc}
    \toprule                                       
    \multirow{2}{*}{Method} &  & \multicolumn{4}{c}{\blue{NAB}} &  & \multicolumn{4}{c}{UCR} &  & \multicolumn{4}{c}{MBA}\tabularnewline
    \cmidrule{3-16} 
     &  & P & R & AUC & F1 &  & P & R & AUC & F1 &  & P & R & AUC & F1\tabularnewline
    \midrule 
    MERLIN &  & 0.8013 & 0.7262 & 0.8414 & 0.7619 &  & 0.7542 & 0.8018 & 0.8984 & 0.7773 &  & 0.9846 & 0.4913 & 0.7828 & 0.6555\tabularnewline
    LSTM-NDT &  & 0.6400 & 0.6667 & 0.8322 & 0.6531 &  & 0.5231 & 0.8294 & 0.9781 & 0.6416 &  & 0.9207 & 0.9718 & 0.9780 & 0.9456\tabularnewline
    DAGMM &  & 0.7622 & 0.7292 & 0.8572 & 0.7453 &  & 0.5337 & 0.9718 & 0.9916 & 0.6890 &  & 0.9475 & 0.9900 & 0.9858 & 0.9683\tabularnewline
    OmniAnomaly &  & 0.8421 & 0.6667 & 0.8330 & 0.7442 &  & 0.8346 & 0.9999 & 0.9981 & 0.9098 &  & 0.8561 & 1.0000 & 0.9570 & 0.9225\tabularnewline
    MSCRED &  & 0.8522 & 0.6700 & 0.8401 & 0.7502 &  & 0.5441 & 0.9718 & 0.9920 & 0.6976 &  & 0.9272 & 1.0000 & 0.9799 & 0.9623\tabularnewline
    MAD-GAN &  & 0.8666 & 0.7012 & 0.8478 & 0.7752 &  & 0.8538 & 0.9891 & 0.9984 & 0.9165 &  & 0.9396 & 1.0000 & 0.9836 & 0.9689\tabularnewline
    USAD &  & 0.8421 & 0.6667 & 0.8330 & 0.7442 &  & 0.8952 & 1.0000 & 0.9989 & 0.9447 &  & 0.8953 & 0.9989 & 0.9701 & 0.9443\tabularnewline
    MTAD-GAT &  & 0.8421 & 0.7272 & 0.8221 & 0.7804 &  & 0.7812 & 0.9972 & 0.9978 & 0.8761 &  & 0.9018 & 1.0000 & 0.9721 & 0.9484\tabularnewline
    CAE-M &  & 0.7918 & 0.8019 & 0.8019 & 0.7968 &  & 0.6981 & 1.0000 & 0.9957 & 0.8222 &  & 0.8442 & 0.9997 & 0.9661 & 0.9154\tabularnewline
    GDN &  & 0.8129 & 0.7872 & 0.8542 & 0.7998 &  & 0.6894 & 0.9988 & 0.9959 & 0.8158 &  & 0.8832 & 0.9892 & 0.9528 & 0.9332\tabularnewline
    \textbf{TranAD} &  & 0.8889 & 0.9892 & \textbf{0.9541} & \textbf{0.9364} &  & 0.9407 & 1.0000 & \textbf{0.9994} & \textbf{0.9694} &  & 0.9569 & 1.0000 & \textbf{0.9885} & \textbf{0.9780}\tabularnewline
    \midrule 
    \multirow{2}{*}{Method} &  & \multicolumn{4}{c}{SMAP} &  & \multicolumn{4}{c}{MSL} &  & \multicolumn{4}{c}{SWaT}\tabularnewline
    \cmidrule{3-16} 
    &  & P & R & AUC & F1 &  & P & R & AUC & F1 &  & P & R & AUC & F1\tabularnewline    
    \midrule 
    MERLIN &  & 0.1577 & 0.9999 & 0.7426 & 0.2725 &  & 0.2613 & 0.4645 & 0.6281 & 0.3345 &  & 0.6560 & 0.2547 & 0.6175 & 0.3669\tabularnewline
    LSTM-NDT &  & 0.8523 & 0.7326 & 0.8602 & 0.7879 &  & 0.6288 & 1.0000 & 0.9532 & 0.7721 &  & 0.7778 & 0.5109 & 0.7140 & 0.6167\tabularnewline
    DAGMM &  & 0.8069 & 0.9891 & 0.9885 & 0.8888 &  & 0.7363 & 1.0000 & 0.9716 & 0.8482 &  & 0.9933 & 0.6879 & 0.8436 & 0.8128\tabularnewline
    OmniAnomaly &  & 0.8130 & 0.9419 & 0.9889 & 0.8728 &  & 0.7848 & 0.9924 & 0.9782 & 0.8765 &  & 0.9782 & 0.6957 & 0.8467 & 0.8131\tabularnewline
    MSCRED &  & 0.8175 & 0.9216 & 0.9821 & 0.8664 &  & 0.8912 & 0.9862 & 0.9807 & 0.9363 &  & 0.9992 & 0.6770 & 0.8433 & 0.8072\tabularnewline
    MAD-GAN &  & 0.8157 & 0.9216 & 0.9891 & 0.8654 &  & 0.8516 & 0.9930 & 0.9862 & 0.9169 &  & 0.9593 & 0.6957 & 0.8463 & 0.8065\tabularnewline
    USAD &  & 0.7480 & 0.9627 & 0.9890 & 0.8419 &  & 0.7949 & 0.9912 & 0.9795 & 0.8822 &  & 0.9977 & 0.6879 & 0.8460 & 0.8143\tabularnewline
    MTAD-GAT &  & 0.7991 & 0.9991 & 0.9844 & 0.8880 &  & 0.7917 & 0.9824 & 0.9899 & 0.8768 &  & 0.9718 & 0.6957 & 0.8464 & 0.8109\tabularnewline
    CAE-M &  & 0.8193 & 0.9567 & 0.9901 & 0.8827 &  & 0.7751 & 1.0000 & 0.9903 & 0.8733 &  & 0.9697 & 0.6957 & 0.8464 & 0.8101\tabularnewline
    GDN &  & 0.7480 & 0.9891 & 0.9864 & 0.8518 &  & 0.9308 & 0.9892 & 0.9814 & \textbf{0.9591} &  & 0.9697 & 0.6957 & 0.8462 & 0.8101\tabularnewline
    \textbf{TranAD} &  & 0.8043 & 0.9999 & \textbf{0.9921} & \textbf{0.8915} &  & 0.9038 & 0.9999 & \textbf{0.9916} & 0.9494 &  & 0.9760 & 0.6997 & \textbf{0.8491} & \textbf{0.8151}\tabularnewline
    \midrule 
    \multirow{2}{*}{Method} &  & \multicolumn{4}{c}{WADI} &  & \multicolumn{4}{c}{SMD} &  & \multicolumn{4}{c}{MSDS}\tabularnewline
    \cmidrule{3-16} 
     &  & P & R & AUC & F1 &  & P & R & AUC & F1 &  & P & R & AUC & F1\tabularnewline
    \midrule 
    MERLIN &  & 0.0636 & 0.7669 & 0.5912 & 0.1174 &  & 0.2871 & 0.5804 & 0.7158 & 0.3842 &  & 0.7254 & 0.3110 & 0.5022 & 0.4353\tabularnewline
    LSTM-NDT &  & 0.0138 & 0.7823 & 0.6721 & 0.0271 &  & 0.9736 & 0.8440 & 0.9671 & 0.9042 &  & 0.9999 & 0.8012 & 0.8013 & 0.8896\tabularnewline
    DAGMM &  & 0.0760 & 0.9981 & 0.8563 & 0.1412 &  & 0.9103 & 0.9914 & 0.9954 & 0.9491 &  & 0.9891 & 0.8026 & 0.9013 & 0.8861\tabularnewline
    OmniAnomaly &  & 0.3158 & 0.6541 & 0.8198 & 0.4260 &  & 0.8881 & 0.9985 & 0.9946 & 0.9401 &  & 1.0000 & 0.7964 & 0.8982 & 0.8867\tabularnewline
    MSCRED &  & 0.2513 & 0.7319 & 0.8412 & 0.3741 &  & 0.7276 & 0.9974 & 0.9921 & 0.8414 &  & 1.0000 & 0.7983 & 0.8943 & 0.8878\tabularnewline
    MAD-GAN &  & 0.2233 & 0.9124 & 0.8026 & 0.3588 &  & 0.9991 & 0.8440 & 0.9933 & 0.9150 &  & 0.9982 & 0.6107 & 0.8054 & 0.7578\tabularnewline
    USAD &  & 0.1873 & 0.8296 & 0.8723 & 0.3056 &  & 0.9060 & 0.9974 & 0.9933 & 0.9495 &  & 0.9912 & 0.7959 & 0.8979 & 0.8829\tabularnewline
    MTAD-GAT &  & 0.2818 & 0.8012 & 0.8821 & 0.4169 &  & 0.8210 & 0.9215 & 0.9921 & 0.8683 &  & 0.9919 & 0.7964 & 0.8982 & 0.8835\tabularnewline
    CAE-M &  & 0.2782 & 0.7918 & 0.8728 & 0.4117 &  & 0.9082 & 0.9671 & 0.9783 & 0.9367 &  & 0.9908 & 0.8439 & 0.9013 & 0.9115\tabularnewline
    GDN &  & 0.2912 & 0.7931 & 0.8777 & 0.4260 &  & 0.7170 & 0.9974 & 0.9924 & 0.8342 &  & 0.9989 & 0.8026 & \textbf{0.9105} & 0.8900\tabularnewline
    \textbf{TranAD} &  & 0.3529 & 0.8296 & \textbf{0.8968} & \textbf{0.4951} &  & 0.9262 & 0.9974 & \textbf{0.9974} & \textbf{0.9605} &  & 0.9999 & 0.8626 & 0.9013 & \textbf{0.9262}\tabularnewline
    \bottomrule 
    \end{tabular}
    \label{tab:detection}
\end{table*}

\subsection{Evaluation Metrics}

\subsubsection{Anomaly Detection}
We use precision, recall, area under the receiver operating characteristic curve (ROC/AUC) and F1 score to evaluate the detection performance of all models~\cite{gdn, mtad_gat}. We also measure the AUC and F1 scores by training all models with $20\%$ of the training data (again using the 80:20 split for validation dataset and the rest as the test set), and call these AUC* and F1* respectively, to measure the performance of the models with limited data. We train on the five sets of 20\% training data and report average results for statistical significance.

\subsubsection{Anomaly Diagnosis}
We use commonly used metrics to measure the diagnosis performance of all models~\cite{mtad_gat}. $\mathrm{HitRate@P\%}$ is the measure of how many ground truth dimensions have been included in the top candidates predicted by the model~\cite{omnianomaly}. $P\%$ is the percentage of the ground truth dimensions for each timestamp, which we use to consider the top predicted candidates. For instance, if at timestamp $t$, if 2 dimensions are labeled anomalous in the ground truth, $\mathrm{HitRate@100\%}$ would consider top 2 dimensions and $\mathrm{HitRate@150\%}$ would consider 3 dimensions (100 and 150 are chosen based on prior work~\cite{mtad_gat}). We also measure the Normalized Discounted Cumulative Gain (NDCG)~\cite{jarvelin2002cumulated}. $\mathrm{NDCG@P\%}$ considers the same number of top predicted candidates as $\mathrm{HitRate@P\%}$.

\begin{table}[]
    \centering 
    \caption{Performance comparison of TranAD with baseline methods with 20\% of the training dataset.  AUC*: AUC with 20\% training data, F1*: F1 score with 20\% training data. The best F1* and AUC* scores are highlighted in bold.} 
    \resizebox{\linewidth}{!}{
    \begin{tabular}{@{}lcccccc@{}}
    \toprule 
    \multirow{2}{*}{Method} & \multicolumn{2}{c}{NAB} & \multicolumn{2}{c}{UCR} & \multicolumn{2}{c}{MBA}\tabularnewline
    \cmidrule{2-7} 
     & AUC{*} & F1{*} & AUC{*} & F1{*} & AUC{*} & F1{*}\tabularnewline
    \midrule 
    MERLIN & \blue{0.8029} & \blue{0.7619} & 0.8984 & 0.7773 & 0.7828 & 0.6555\tabularnewline
    LSTM-NDT & \blue{0.8013} & \blue{0.6212} & 0.8913 & 0.5198 & 0.9617 & 0.9282\tabularnewline
    DAGMM & \blue{0.7827} & \blue{0.6125} & 0.9812 & 0.5718 & 0.9671 & 0.9396\tabularnewline
    OmniAnomaly & \blue{0.8129} & \blue{0.6713} & 0.9728 & 0.7918 & 0.9407 & 0.9217\tabularnewline
    MSCRED & \blue{0.8299} & \blue{0.7013} & 0.9637 & 0.4929 & 0.9499 & 0.9108\tabularnewline
    MAD-GAN & \blue{0.8194} & \blue{0.7109} & 0.9959 & 0.8216 & 0.9550 & 0.9192\tabularnewline
    USAD & \blue{0.7267} & \blue{0.6781} & 0.9967 & 0.8538 & 0.9697 & 0.9425\tabularnewline
    MTAD-GAT & \blue{0.6956} & \blue{0.7013} & 0.9974 & 0.8671 & 0.9688 & 0.9425\tabularnewline
    CAE-M & \blue{0.7312} & \blue{0.7126} & 0.9926 & 0.7525 & 0.9616 & 0.9002\tabularnewline
    GDN & \blue{0.8299} & \blue{0.7013} & 0.9937 & 0.8029 & 0.9671 & 0.9316\tabularnewline
    \textbf{TranAD} & \textbf{0.9217} & \textbf{0.8421} & \textbf{0.9989} & \textbf{0.9399} & \textbf{0.9718} & \textbf{0.9617}\tabularnewline
    \midrule 
    \multirow{2}{*}{Method} & \multicolumn{2}{c}{SMAP} & \multicolumn{2}{c}{MSL} & \multicolumn{2}{c}{SWaT}\tabularnewline
    \cmidrule{2-7} 
     & AUC{*} & F1{*} & AUC{*} & F1{*} & AUC{*} & F1{*}\tabularnewline
    \midrule 
    MERLIN & 0.7426 & 0.2725 & 0.6281 & 0.3345 & 0.6175 & 0.3669\tabularnewline
    LSTM-NDT & 0.7007 & 0.5418 & 0.9520 & 0.7608 & 0.6690 & 0.4145\tabularnewline
    DAGMM & 0.9881 & 0.8369 & 0.9606 & 0.8010 & 0.8421 & 0.8001\tabularnewline
    OmniAnomaly & 0.9879 & 0.8131 & 0.9703 & 0.8424 & 0.8319 & 0.7433\tabularnewline
    MSCRED & 0.9811 & 0.8050 & 0.9797 & 0.8232 & 0.8385 & 0.7922\tabularnewline
    MAD-GAN & 0.9877 & 0.8468 & 0.9649 & 0.8190 & 0.8456 & 0.8012\tabularnewline
    USAD & 0.9883 & 0.8379 & 0.9649 & 0.8190 & 0.8438 & 0.8087\tabularnewline
    MTAD-GAT & 0.9814 & 0.8225 & 0.9782 & 0.8024 & \textbf{0.8459} & 0.8079\tabularnewline
    CAE-M & \textbf{0.9892} & 0.8312 & 0.9836 & 0.7303 & 0.8458 & 0.7841\tabularnewline
    GDN & 0.9887 & 0.8411 & 0.9414 & 0.8959 & 0.8390 & 0.8072\tabularnewline
    \textbf{TranAD} & 0.9885 & \textbf{0.8889} & \textbf{0.9857} & \textbf{0.9172} & 0.8438 & \textbf{0.8094}\tabularnewline
    \midrule 
    \multirow{2}{*}{Method} & \multicolumn{2}{c}{WADI} & \multicolumn{2}{c}{SMD} & \multicolumn{2}{c}{MSDS}\tabularnewline
    \cmidrule{2-7} 
     & AUC{*} & F1{*} & AUC{*} & F1{*} & AUC{*} & F1{*}\tabularnewline
    \midrule 
    MERLIN & 0.5912 & \textbf{0.1174} & 0.7158 & 0.3842 & 0.5022 & 0.4353\tabularnewline
    LSTM-NDT & 0.6637 & 0.0000 & 0.9563 & 0.6754 & 0.7813 & 0.7912\tabularnewline
    DAGMM & 0.6497 & 0.0630 & 0.9845 & 0.8986 & 0.7763 & 0.8389\tabularnewline
    OmniAnomaly & \textbf{0.7913} & 0.1017 & 0.9859 & 0.9352 & 0.5613 & 0.8389\tabularnewline
    MSCRED & 0.6029 & 0.0413 & 0.9768 & 0.8004 & 0.7716 & 0.8283\tabularnewline
    MAD-GAN & 0.5383 & 0.0937 & 0.8635 & 0.9318 & 0.5002 & 0.7390\tabularnewline
    USAD & 0.7011 & 0.0733 & 0.9854 & 0.9213 & 0.7613 & 0.8389\tabularnewline
    MTAD-GAT & 0.6267 & 0.0520 & 0.9798 & 0.6661 & 0.6122 & 0.8248\tabularnewline
    CAE-M & 0.6109 & 0.0781 & 0.9569 & 0.9318 & 0.6001 & 0.8389\tabularnewline
    GDN & 0.6121 & 0.0412 & 0.9811 & 0.7107 & 0.6819 & 0.8389\tabularnewline
    \textbf{TranAD} & 0.7688 & 0.0649 & \textbf{0.9869} & \textbf{0.9478} & \textbf{0.8113} & \textbf{0.8391}\tabularnewline
    \bottomrule 
    \end{tabular}}
    \label{tab:detection20}
\end{table}

\subsection{Results}
\textbf{Anomaly Detection.} Tables~\ref{tab:detection} \blue{and~\ref{tab:detection20}} provide the precision, recall, AUC, F1, AUC* and F1* scores for TranAD and baseline models for all datasets. On average, the F1 score of the TranAD model is \blue{$0.8802$} and F1* is \blue{$0.8012$}. TranAD outperforms the baselines (in terms of F1 score) for all datasets except MSL when we consider the complete dataset for model training. TranAD also outperforms baselines for all datasets except the WADI dataset with $20\%$ of the dataset used for training (F1* score). For MSL, the GDN model has the highest F1 score ($0.9591$) and for the WADI dataset, OmniAnomaly has the highest F1* score ($0.1017$). Similarly, TranAD outperforms baselines in terms of AUC scores for all datasets except MSDS, where GDN has the highest AUC ($0.9105$). All models perform relatively poorly on WADI due to its large-scale in terms of sequence lengths and data modality. Specifically, TranAD achieves improvement of up to \blue{$17.06\%$} in F1 score, $14.64\%$ in F1* score, \blue{$11.69\%$} in AUC and $11.06\%$ in AUC* scores over the state-of-the-art baseline models.


The MERLIN baseline is a parameter free approach that does not require any training data; hence, we report F1* and AUC* as F1 and AUC scores, respectively. \blue{MERLIN performs relatively well only on the univariate datasets, \textit{i.e.} NAB and UCR, and is unable to scale effectively to multivariate data in our traces.} The baseline method LSTM-NDT has a good performance on MSL and SMD, but performs poorly on other datasets. This is due to its sensitivity to different scenarios and poor efficiency of the NDT thresholding method~\cite{mtad_gat}. \blue{The POT technique used in TranAD and other models like OmniAnomaly helps set more accurate threshold values by also considering the localized peak values in the data sequence.} DAGMM model performs very well for short datasets like UCR, \blue{NAB}, MBA and SMAP, but its scores drop significantly for other datasets with longer sequences. This is because it does not map the temporal information explicitly as it does not use sequence windows but only a single GRU model. \blue{The window encoder in TranAD, with the encoding of the complete sequence as a self-condition, allows it to perform better even with long high-dimensional sequences.} The OmniAnomaly, CAE-M and MSCRED models use sequential observations as input, allowing these methods to retain the temporal information. Such methods perform reconstruction regardless of anomalous data, which prevents them from detecting anomalies close to the normal trends~\cite{audibert2020usad}. \blue{TranAD tackles this by using adversarial training to amplify errors. Hence, in datasets like SMD, where anomalous data is not very far from normal data, it can detect even mild anomalies.}

\begin{figure}[!t]
    \centering 
    \includegraphics[width=\linewidth]{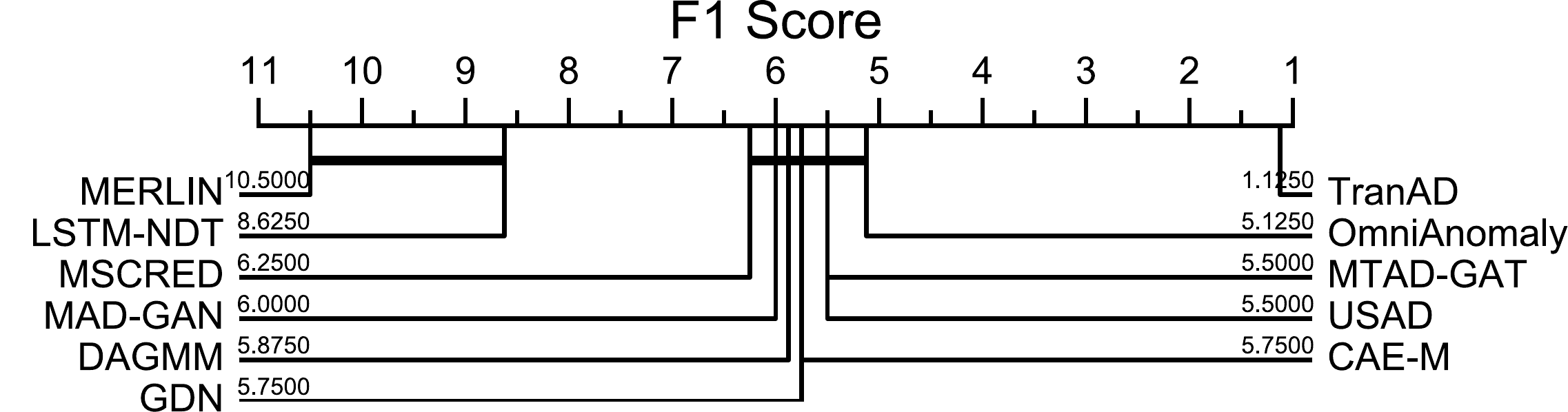}
    \includegraphics[width=\linewidth]{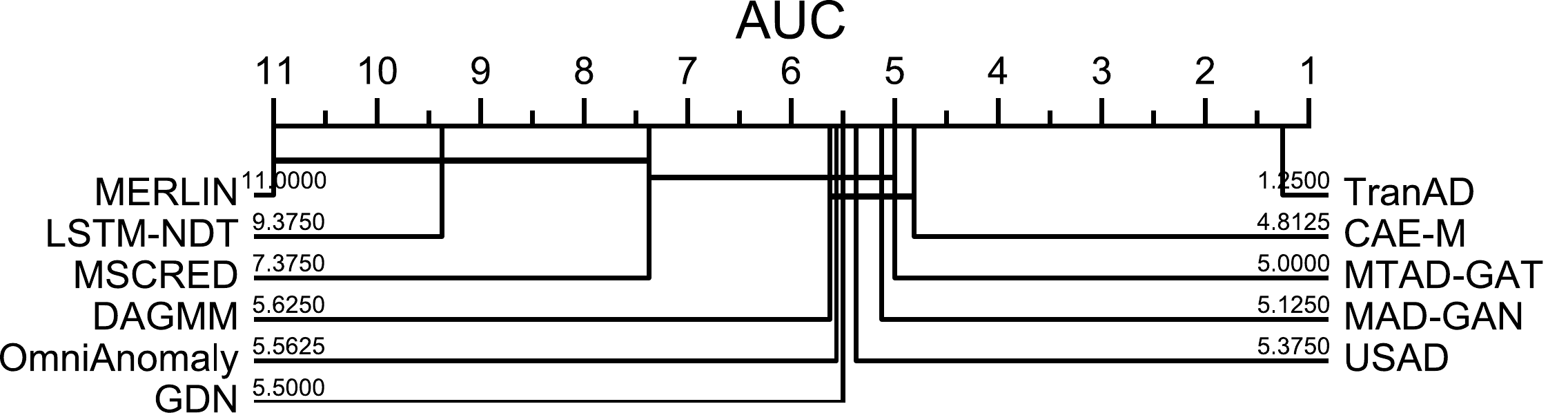}
    \caption{Critical difference diagrams for F1 and AUC scores using the Wilkoxon pairwised signed rank test (with $\alpha=0.05$) on all datasets. Rightmost methods are ranked higher.}
    \label{fig:cd}
\end{figure}
\begin{figure}[]
    \centering 
    \includegraphics[width=\linewidth]{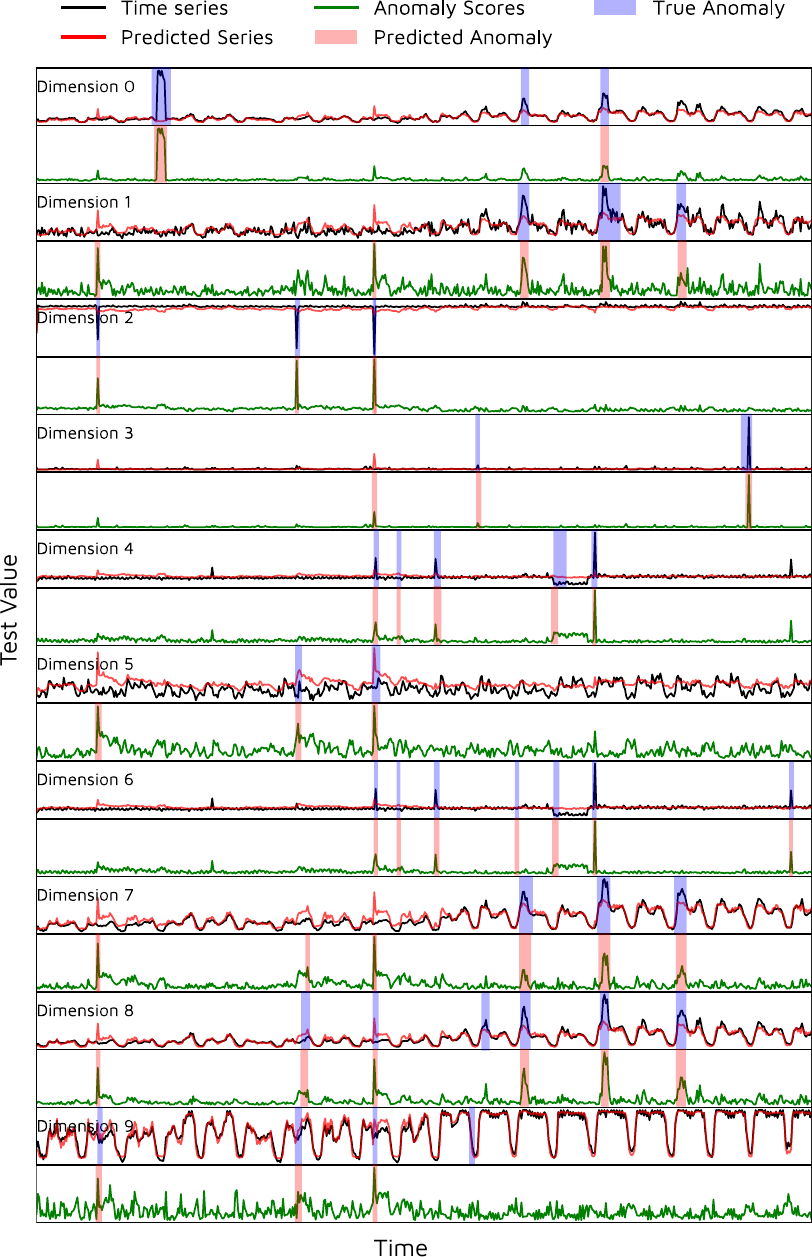}
    \caption{Predicted and Ground Truth labels for the MSDS test set using the TranAD model.}
    \label{fig:prediction}
\end{figure}
\begin{table*}[]
    \centering 
    \caption{Diagnosis Performance.}
    \begin{tabular}{llccccccccc}
    \toprule 
    \multirow{2}{*}{Method} &  & \multicolumn{4}{c}{SMD} &  & \multicolumn{4}{c}{MSDS}\tabularnewline
    \cmidrule{3-11}
     &  & H@100\% & H@150\% & N@100\% & N@150\% &  & H@100\% & H@150\% & N@100\% & N@150\%\tabularnewline
    \midrule 
    MERLIN &  & \textbf{0.5907} & 0.6177 & 0.4150 & 0.4912 &  & 0.3816 & 0.5626 & 0.3010 & 0.3947\tabularnewline
    LSTM-NDT &  & 0.3808 & 0.5225 & 0.3603 & 0.4451 &  & 0.1504 & 0.2959 & 0.1124 & 0.1993\tabularnewline
    DAGMM &  & 0.4927 & 0.6091 & 0.5169 & 0.5845 &  & 0.2617 & 0.4333 & 0.3153 & 0.4154\tabularnewline
    OmniAnomaly &  & 0.4567 & 0.5652 & 0.4545 & 0.5125 &  & 0.2839 & 0.4365 & 0.3338 & 0.4231\tabularnewline
    MSCRED &  & 0.4272 & 0.5180 & 0.4609 & 0.5164 &  & 0.2322 & 0.3469 & 0.2297 & 0.2962\tabularnewline
    MAD-GAN &  & 0.4630 & 0.5785 & 0.4681 & 0.5522 &  & 0.3856 & 0.5589 & 0.4277 & 0.5292\tabularnewline
    USAD &  & 0.4925 & 0.6055 & 0.5179 & 0.5781 &  & 0.3095 & 0.4769 & 0.3534 & 0.4515\tabularnewline
    MTAD-GAT &  & 0.3493 & 0.4777 & 0.3759 & 0.4530 &  & \textbf{0.5812} & 0.5885 & 0.5926 & 0.6522\tabularnewline
    CAE-M &  & 0.4707 & 0.5878 & \textbf{0.5474} & \textbf{0.6178} &  & 0.2530 & 0.4171 & 0.2047 & 0.3010\tabularnewline
    GDN &  & 0.3143 & 0.4386 & 0.2980 & 0.3724 &  & 0.2276 & 0.3382 & 0.2921 & 0.3570\tabularnewline
    \textbf{TranAD} &  & 0.4981 & \textbf{0.6401} & 0.4941 & \textbf{0.6178} &  & 0.4630 & \textbf{0.7533} & \textbf{0.5981} & \textbf{0.6963}\tabularnewline
    \bottomrule 
    \end{tabular}
    \label{tab:diagnosis}
\end{table*}

\begin{table*}[]
    \centering 
    \caption{Comparison of training times in seconds per epoch.}
    \begin{tabular}{lllcccccccc}
    \toprule 
    Method &  & NAB & UCR & MBA & SMAP & MSL & SWaT & WADI & SMD & MSDS\tabularnewline
    \midrule 
    MERLIN &  & \blue{3.28} & 4.09 & 20.19 & 6.89 & \textbf{5.12} & 10.12 & 132.69 & 72.32 & 42.22\tabularnewline
    LSTM-NDT &  & \blue{10.64} & 8.71 & 27.80 & 27.62 & 26.24 & 26.43 & 297.12 & 373.14 & 361.12\tabularnewline
    DAGMM &  & \blue{25.38} & 20.78 & 74.62 & 19.05 & 16.41 & 18.51 & 178.17 & 204.36 & 187.54\tabularnewline
    OmniAnomaly &  & \blue{38.27} & 27.96 & 109.86 & 27.05 & 21.31 & 28.39 & 212.99 & 276.97 & 277.10\tabularnewline
    MSCRED &  & \blue{258.86} & 262.45 & 592.13 & 16.13 & 33.47 & 183.67 & 1349.05 & 237.66 & 109.63\tabularnewline
    MAD-GAN &  & \blue{39.80} & 25.71 & 160.29 & 29.49 & 26.27 & 27.79 & 293.60 & 314.82 & 285.25\tabularnewline
    USAD &  & \blue{31.21} & 21.10 & 120.86 & 23.63 & 21.22 & 22.72 & 242.86 & 250.97 & 232.82\tabularnewline
    MTAD-GAT &  & \blue{145.00} & 97.12 & 233.08 & 1015.03 & 1287.42 & 103.92 & 9812.13 & 6564.11 & 1304.09\tabularnewline
    CAE-M &  & \blue{22.48} & 19.42 & 67.44 & 187.35 & 575.96 & 41.25 & 5525.62 & 3102.12 & 552.83\tabularnewline
    GDN &  & \blue{83.84} & 58.78 & 159.01 & 62.33 & 96.71 & 59.40 & 4063.05 & 809.94 & 585.34\tabularnewline
    \textbf{TranAD} &  & \blue{\textbf{1.25}} & \textbf{0.84} & \textbf{4.08} & \textbf{3.55} & 5.27 & \textbf{0.87} & \textbf{115.91} & \textbf{43.56} & \textbf{17.15}\tabularnewline
    \bottomrule 
    \end{tabular}
    \label{tab:overhead}
\end{table*}

Recent models such as USAD, MTAD-GAT and GDN use attention mechanisms to focus on specific modes of the data. Moreover, these models try to capture the long-term trends by adjusting the weights of their neural network and only use a local window as an input for reconstruction. \blue{GDN has slightly higher scores for MSL and MSDS datasets than TranAD due to the scalable graph-based inference over the inter-dimensional data correlations~\cite{gdn}. TranAD does this using self-attention and performs better than GDN overall across all datasets.} The limitation of seeing only a local contextual window prevents methods such as USAD and MTAD-GAT from classifying long-term anomalies (like in SMD or WADI). However, self-conditioning on an embedding of the complete trace with position encoding aids temporal attention, thanks to the transformer architecture in TranAD. This allows TranAD to capture long-term trends more effectively. Further, due to the meta-learning, TranAD also outperforms baselines with limited training data except for OmniAnomaly on the WADI dataset, indicating its high efficacy even with limited data. \blue{OmniAnomaly performs best among all methods on the WADI dataset due to high noise in this dataset and dedicated stochasticity modeling in OmniAnomaly~\cite{omnianomaly}. TranAD is slightly behind this method in terms of F1* and AUC*; however, outperforms it when compared across all datasets and also when given the complete WADI dataset.}

We perform critical difference analysis to assess the significance of the differences among the performance of the models. Figure~\ref{fig:cd} depicts the critical difference diagrams for the F1 and AUC scores based on the Wilcoxon pair-wised signed-rank test (with $\alpha = 0.05$) after rejecting the null hypothesis using the Friedman test on all datasets~\cite{criticaldifference}. TranAD achieves the best rank across all models with a significant statistical difference.

\begin{table}[!t]
    \centering 
    \caption{Ablation Study - F1 and F1* scores for TranAD and its ablated versions.}
    \resizebox{\linewidth}{!}{
    \begin{tabular}{@{}lcccccc@{}}
    \toprule 
    \multirow{2}{*}{Method} & \multicolumn{2}{c}{NAB} & \multicolumn{2}{c}{UCR} & \multicolumn{2}{c}{MBA}\tabularnewline
    \cmidrule{2-7} 
     & F1 & F1{*} & F1 & F1{*} & F1 & F1{*}\tabularnewline
    \midrule 
    TranAD & \textbf{0.9364} & \textbf{0.8421} & \textbf{0.9694} & \textbf{0.9399} & \textbf{0.9780} & \textbf{0.9617}\tabularnewline
    \hline 
    w/o transformer & 0.8850 & 0.8019 & 0.8466 & 0.5495 & 0.9749 & 0.9584\tabularnewline
    w/o self-condition & 0.8887 & 0.8102 & 0.9191 & 0.9028 & 0.9770 & \textbf{0.9617}\tabularnewline
    w/o adversarial training & 0.9012 & 0.8102 & 0.9634 & 0.9289 & 0.9752 & 0.9592\tabularnewline
    w/o MAML & 0.9068 & 0.8210 & 0.9689 & 0.9304 & 0.9756 & \textbf{0.9617}\tabularnewline
    \midrule 
    \multirow{2}{*}{Method} & \multicolumn{2}{c}{SMAP} & \multicolumn{2}{c}{MSL} & \multicolumn{2}{c}{SWaT}\tabularnewline
    \cmidrule{2-7} 
     & F1 & F1{*} & F1 & F1{*} & F1 & F1{*}\tabularnewline
    \midrule 
    TranAD & \textbf{0.8915} & \textbf{0.8889} & \textbf{0.9494} & \textbf{0.9172} & \textbf{0.8151} & \textbf{0.8094}\tabularnewline
    \hline 
    w/o transformer & 0.8643 & 0.8147 & 0.9137 & 0.9037 & 0.8143 & 0.6360\tabularnewline
    w/o self-condition & 0.8894 & 0.8153 & 0.9175 & 0.8913 & 0.7953 & 0.8094\tabularnewline
    w/o adversarial training & 0.8906 & 0.8476 & 0.9455 & \textbf{0.9172} & 0.8028 & 0.7832\tabularnewline
    w/o MAML & \textbf{0.8915} & \textbf{0.8899} & 0.9466 & 0.6732 & 0.8143 & 0.8079\tabularnewline
    \midrule 
    \multirow{2}{*}{Method} & \multicolumn{2}{c}{WADI} & \multicolumn{2}{c}{SMD} & \multicolumn{2}{c}{MSDS}\tabularnewline
    \cmidrule{2-7} 
     & F1 & F1{*} & F1 & F1{*} & F1 & F1{*}\tabularnewline
    \midrule 
    TranAD & \textbf{0.4951} & \textbf{0.0649} & \textbf{0.9605} & \textbf{0.9478} & \textbf{0.9262} & \textbf{0.8391}\tabularnewline
    \hline 
    w/o transformer & 0.2181 & 0.0037 & 0.9071 & 0.9032 & 0.8867 & 0.8389\tabularnewline
    w/o self-condition & 0.3620 & 0.0631 & 0.9502 & 0.8847 & 0.8748 & 0.8214\tabularnewline
    w/o adversarial training & 0.3820 & 0.0621 & 0.9177 & 0.8667 & 0.9181 & 0.8389\tabularnewline
    w/o MAML & 0.4815 & 0.0553 & 0.9433 & 0.8164 & 0.8870 & 0.8389\tabularnewline
    \bottomrule 
    \end{tabular}}
    \label{tab:ablation}
\end{table}

\textbf{Anomaly Diagnosis.} The anomaly diagnosis results in Table~\ref{tab:diagnosis} where H and N correspond to HitRate and NDCG  (with complete data used for model testing). We only present results on the \blue{multivariate} SMD and MSDS datasets for the sake of brevity (TranAD yields better scores for others as well). We also ignore models that do not explicitly output anomaly class outputs for each dimension individually. Multi-head attention in TranAD allows it to attend to multiple modes simultaneously, making it more suitable for more inter-correlated anomalies. This is observed and explained by datasets like MSDS (distributed systems) where anomalous behavior in one mode can lead to a chain of events causing anomalies in other modes (see Figure~\ref{fig:prediction}). TranAD is able to leverage the complete trace information with the local window to aid in pinpointing anomalous behavior to specific modes. The table demonstrates that TranAD is able to detect $46.3\%-75.3\%$ root causes for anomalies. 

Compared to the baseline methods, TranAD is able to improve diagnosis score by up to $6\%$ for SMD and $30\%$ for MSDS. The average improvement in diagnosis scores is $4.25\%$.

\section{Analyses}
\label{sec:analyses}

\subsection{Ablation Analysis}
To study the relative importance of each component of the model, we exclude every major one and observe how it affects the performance in terms of the F1 scores for each dataset. First, we consider the TranAD model without the transformer-based encoder-decoder architecture but instead with a feed-forward network. Second, we consider the model without the self-conditioning, \textit{i.e.}, we fix the focus score $F = \vec{0}$ in phase 2. Third, we study the model without the adversarial loss, \textit{i.e.}, a single-phase inference and only the reconstruction loss for model training. Finally, we consider the model without meta-learning. The results are summarized in Table~\ref{tab:ablation} and provide the following findings:
\begin{itemize}[leftmargin=*]
    \item Replacing the transformer-based encoder-decoder has the highest performance drop of nearly \blue{$11\%$} in terms of the F1 score. This drop is more pronounced for the WADI dataset ($56\%$), demonstrating the need for the attention-based transformer for large-scale datasets.
    \item When we remove the self-conditioning, the average drop in F1 scores is $6\%$, which shows that the focus score aids prediction performance.
    \item Removing the two-phase adversarial training mainly affects SMD and WADI datasets as these traces have a large proportion of mild anomalies and the adversarial loss helps amplify the anomaly scores. The average drop in F1 score, in this case, is $5\%$.
    \item \blue{Not having the meta-learning in the model has little effect to the F1 scores ($\approx\!1\%$); however, it leads to a nearly $12\%$ drop in F1* scores.}
\end{itemize}

\begin{figure*}
    \centering 
    \includegraphics[width=.6\linewidth]{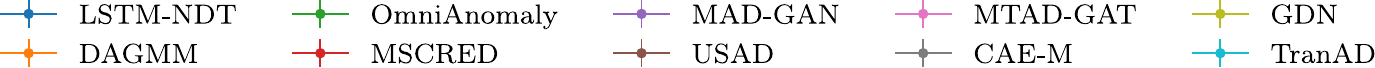}\\
    \subfigure[F1 Score]{
    \includegraphics[width=.305\linewidth]{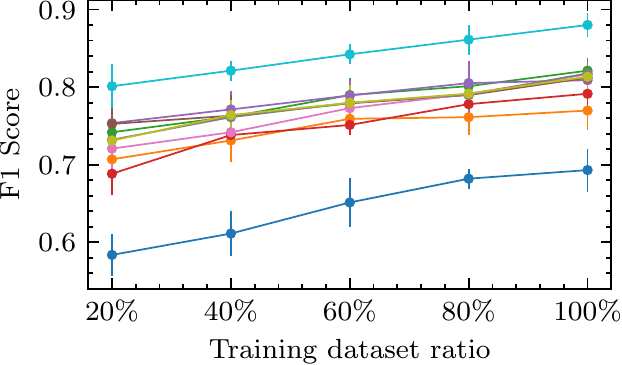}
    \label{fig:m_f1}
    }
    \subfigure[ROC/AUC Score]{
    \includegraphics[width=.305\linewidth]{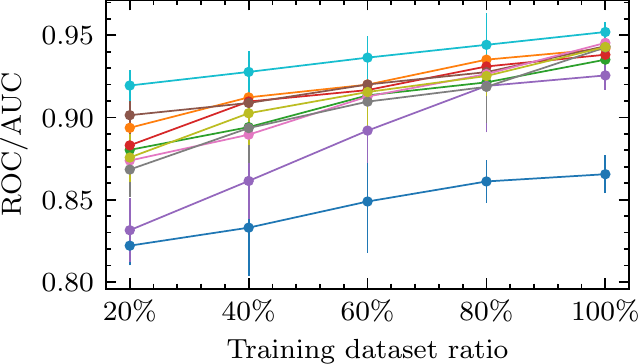}
    \label{fig:m_acu}
    }
    \subfigure[Training Time]{
    \includegraphics[width=.305\linewidth]{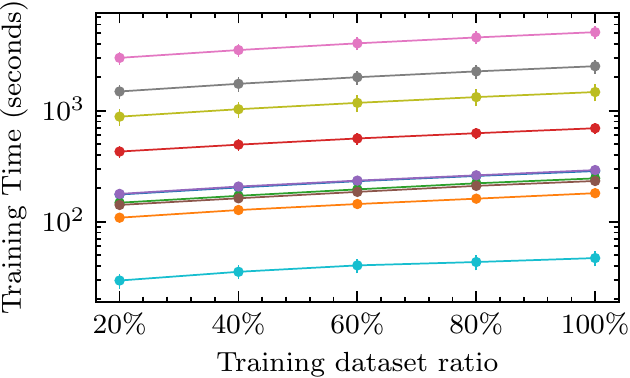}
    \label{fig:m_training}
    }
    \caption{F1 score, ROC/AUC score and training times with dataset size.}
    \label{fig:sen_m}
\end{figure*}
\begin{figure*}
    \centering 
    \includegraphics[width=0.9\linewidth]{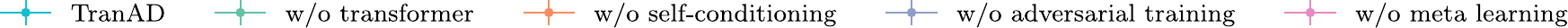}\\
    \subfigure[F1 Score]{
    \includegraphics[width=.305\linewidth]{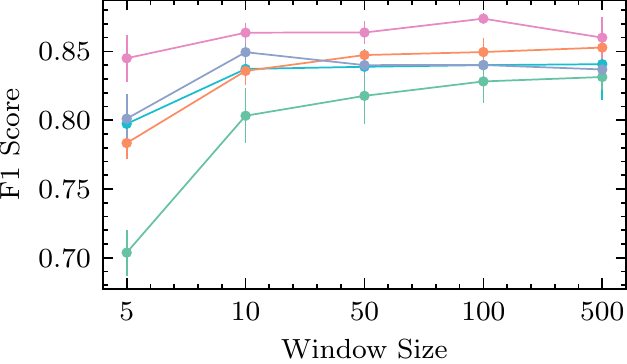}
    \label{fig:ab_f1}
    }
    \subfigure[ROC/AUC Score]{
    \includegraphics[width=.305\linewidth]{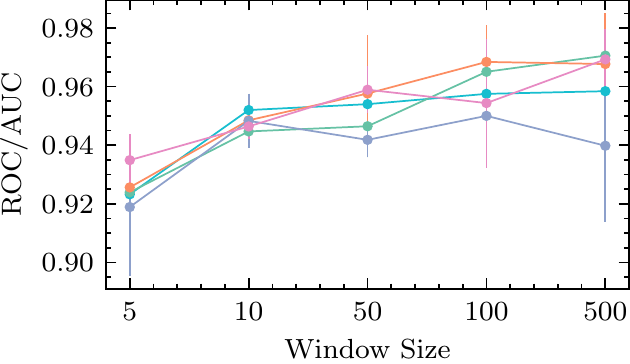}
    \label{fig:ab_auc}
    }
    \subfigure[Training Time]{
    \includegraphics[width=.305\linewidth]{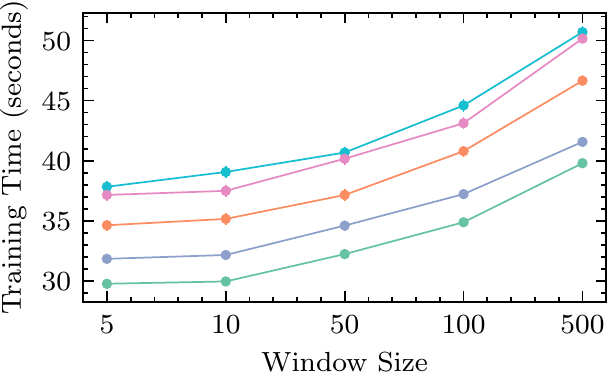}
    \label{fig:ab_training}
    }
    \caption{F1 score, ROC/AUC score and training times with window size.}
    \label{fig:sen_ab}
\end{figure*}

\subsection{Overhead Analysis}
Table~\ref{tab:overhead} shows the average training times for all models on every dataset in seconds per epoch. For comparison, we report the training time for MERLIN as the time it takes to discover discords in the test data. The training time of TranAD is $75\%-99\%$ lower than those of the baseline methods. This clearly shows the advantage of having a transformer model with positional encoding to push the complete sequence as an input instead of sequentially inferring over local windows.

\subsection{Sensitivity Analysis}

\textbf{Sensitivity to the training set size.} Figure~\ref{fig:sen_m} shows the variation of the F1 and AUC scores of all models averaged for all datasets, and the training time with the ratio of the training data used for model training, ranging from $20\%$ to $100\%$. We do not report sensitivity results on MERLIN as it does not use training data. Other deep learning based reconstruction models are given the same randomly sampled subsequence of $20\%$ to $100\%$ size as that of the training data. We report $90\%$ confidence bounds in Figure~\ref{fig:sen_m}. Clearly, as dataset size increases, the prediction performance improves and the training time increases. We observe that for every ratio, the TranAD model has a higher F1 score and a lower training time.  

\textbf{Sensitivity to the window size.} We also show the performance of the TranAD model and its variants with different window sizes in Figure~\ref{fig:sen_ab}. The window size has an impact both on the anomaly detection scores and training times. TranAD can detect anomalies faster when we have smaller windows since the inference time is lower for smaller inputs. If the window is too small, it does not represent the local contextual information well. However, if the window is too large, short anomalies may be hidden in a large number of datapoints in such a window (see the drop in F1 score for some models). A window size of 10 gives a reasonable balance between the F1 score and training times and hence is used in our experiments. 

\begin{table}[]
    \centering \renewcommand*{\arraystretch}{1.05}
    \caption{Comparison between original MERLIN implementation (MATLAB) with ours (Python).}
    \label{tab:merlin_impl}
\begin{tabular}{llccc}
\toprule 
Benchmark & Metric & Original & Ours & Deviation\tabularnewline
\midrule
\multirow{5}{*}{NAB} & P & 0.7828 & 0.8013 & 0.0236\tabularnewline
 & R & 0.6981 & 0.7262 & 0.0402\tabularnewline
 & AUC & 0.8201 & 0.8414 & 0.0259\tabularnewline
 & F1 & 0.7380 & 0.7619 & 0.0323\tabularnewline
 & Time & 69.2190 & 3.2800 & -0.9526\tabularnewline
 \midrule
\multirow{5}{*}{UCR} & P & 0.7328 & 0.7542 & 0.0292\tabularnewline
 & R & 0.7813 & 0.8018 & 0.0263\tabularnewline
 & AUC & 0.8562 & 0.8984 & 0.0493\tabularnewline
 & F1 & 0.7563 & 0.7542 & -0.0027\tabularnewline
 & Time & 67.1289 & 4.0900 & -0.9391\tabularnewline
 \midrule
\multirow{5}{*}{MBA} & P & 0.9846 & 0.9846 & 0.0000\tabularnewline
 & R & 0.4827 & 0.4913 & 0.0178\tabularnewline
 & AUC & 0.7627 & 0.7827 & 0.0262\tabularnewline
 & F1 & 0.6478 & 0.6555 & 0.0119\tabularnewline
 & Time & 127.2100 & 20.1900 & -0.8413\tabularnewline
 \midrule
\multirow{5}{*}{SMAP} & P & 0.5494 & 0.6560 & 0.1941\tabularnewline
 & R & 0.2619 & 0.2547 & -0.0274\tabularnewline
 & AUC & 0.6026 & 0.6175 & 0.0248\tabularnewline
 & F1 & 0.3547 & 0.3669 & 0.0345\tabularnewline
 & Time & 105.7820 & 6.8900 & -0.9349\tabularnewline
 \midrule
\multirow{5}{*}{MSL} & P & 0.2827 & 0.2613 & -0.0756\tabularnewline
 & R & 0.4177 & 0.4645 & 0.1119\tabularnewline
 & AUC & 0.5984 & 0.6281 & 0.0497\tabularnewline
 & F1 & 0.3372 & 0.3345 & -0.0081\tabularnewline
 & Time & 167.1230 & 5.1200 & -0.9694\tabularnewline
 \midrule
\multirow{5}{*}{SWaT} & P & 0.6872 & 0.6560 & -0.0454\tabularnewline
 & R & 0.2562 & 0.2547 & -0.0059\tabularnewline
 & AUC & 0.6281 & 0.6175 & -0.0169\tabularnewline
 & F1 & 0.3733 & 0.3669 & -0.0170\tabularnewline
 & Time & 782.1282 & 10.1200 & -0.9871\tabularnewline
 \midrule
\multirow{5}{*}{WADI} & P & 0.0687 & 0.0636 & -0.0745\tabularnewline
 & R & 0.6727 & 0.7669 & 0.1400\tabularnewline
 & AUC & 0.5821 & 0.5912 & 0.0156\tabularnewline
 & F1 & 0.1247 & 0.1174 & -0.0585\tabularnewline
 & Time & 1029.2100 & 132.6900 & -0.8711\tabularnewline
 \midrule
\multirow{5}{*}{SMD} & P & 0.3282 & 0.3287 & 0.0016\tabularnewline
 & R & 0.5793 & 0.5804 & 0.0019\tabularnewline
 & AUC & 0.6918 & 0.7158 & 0.0347\tabularnewline
 & F1 & 0.4190 & 0.3842 & -0.0831\tabularnewline
 & Time & 762.2189 & 72.3200 & -0.9051\tabularnewline
 \midrule
\multirow{5}{*}{MSDS} & P & 0.6981 & 0.7254 & 0.0391\tabularnewline
 & R & 0.3128 & 0.3110 & -0.0058\tabularnewline
 & AUC & 0.5042 & 0.5022 & -0.0040\tabularnewline
 & F1 & 0.4320 & 0.4353 & 0.0076\tabularnewline
 & Time & 892.2890 & 42.2200 & -0.9527\tabularnewline
\bottomrule
\end{tabular}
\end{table}

\section{Conclusions}
\label{sec:conclusion}
\noindent
We present a transformer based anomaly detection model (TranAD) that can detect and diagnose anomalies for multivariate time-series data. The transformer based encoder-decoder allows quick model training and high detection performance for a variety of datasets considered in this work. TranAD leverages self-conditioning and adversarial training to amplify errors and gain training stability. Moreover, meta-learning allows it to be able to identify data trends even with limited data. Specifically, TranAD achieves an improvement of \blue{$17\%$} and \blue{11\%} for F1 score on complete and limited training data, respectively. It is also able to correctly identify root causes for up to $75\%$ of the detected anomalies, higher than the state-of-the-art models. It is able to achieve this with up to $99\%$ lower training times compared to the baseline methods. This makes TranAD an ideal choice for modern industrial systems where accurate and quick anomaly predictions are required.

For the future, we propose to extend the method with other transformer models like bidirectional neural networks to allow model generalization to diverse temporal trends in data. \blue{We also wish to explore the direction of applying cost-benefit analysis for each model component based on the deployment setting to avoid expensive computation}. 

\section*{Software Availability}
\noindent
The code and relevant training scripts are made publicly available on GitHub under BSD-3 licence at \url{https://github.com/imperial-qore/TranAD}.

\begin{acks}
Shreshth Tuli is supported by the President's PhD scholarship at Imperial College London. We thank Kate Highnam for constructive comments on improving the manuscript writing. We thank the providers of all datasets used in this work.
\end{acks}

\appendix

\section{MERLIN Implementation}
\label{app:merlin}

\red{The comparisons shown in Section~\ref{sec:experiments} use a custom implementation of the MERLIN baseline in Python. Table~\ref{tab:merlin_impl} compares the the original MERLIN implementation in MATLAB\footnote{MERLIN: \url{https://sites.google.com/view/merlin-find-anomalies/documentation} [Accessed: 30 January 2022, Last Updated: 23 June 2020].} with our own Python based implementation. As the original code does not present how to select the hyperparameters, \textit{i.e.}, \texttt{MinL} and \texttt{MaxL}, corresponding to the discord lengths, we use grid-search to find optimal parameters by maximizing F1 score. The results in Table~\ref{tab:merlin_impl} use the hyperparameter values as (\texttt{MinL}, \texttt{MaxL}) = \{(10, 40), (50, 60), (60, 100), (70, 100), (30, 60), (10, 20), (60, 100), (100, 140), (5, 10)\} for the datasets NAB, UCR, MBA, SMAP, MSL, SWaT, WADI, SMD and MSDS respectively.}

\red{For every benchmark, we find the metrics, \textit{i.e.}, Precision (P), Recall (R), AUC of the ROC curve (AUC), F1 score (F1), and training time (Time) for the MATLAB code ($x$) and our implementation ($y$). We calculate the deviation as $(y-x)/x$. The table shows that our implementation gives scores very close to the original implementation, with training time gains by up to 96\%.}

\balance
\bibliographystyle{ACM-Reference-Format}
\bibliography{references}

\end{document}